\documentclass[lettersize,journal]{IEEEtran}
\usepackage{amsmath,amsfonts}
\usepackage{algorithmic}
\usepackage{algorithm}
\usepackage{array}
\usepackage[caption=false,font=normalsize,labelfont=sf,textfont=sf]{subfig}
\usepackage{textcomp}
\usepackage{stfloats}
\usepackage{url}
\usepackage{verbatim}
\usepackage{graphicx}
\usepackage{cite}

\usepackage{soul}

\usepackage{bm}
\usepackage[utf8]{inputenc} % allow utf-8 input
\usepackage[T1]{fontenc}    % use 8-bit T1 fonts
\usepackage{url}            % simple URL typesetting
\usepackage{booktabs}       % professional-quality tables
\usepackage{nicefrac}       % compact symbols for 1/2, etc.
\usepackage{microtype}      % microtypography
\usepackage{xcolor}         % colors
\usepackage{graphicx}
\usepackage{dutchcal}
\usepackage{caption}
\newcommand{\cmark}{\ding{51}}%

\usepackage{multirow}
\usepackage{booktabs}

\usepackage{siunitx} 
\usepackage{makecell}
\usepackage{pifont} 
\usepackage{wrapfig}
\usepackage{placeins}

\usepackage{enumitem} 
\setlist[enumerate]{leftmargin=*}
\usepackage{colortbl}
\definecolor{mygray}{gray}{.9}
\definecolor{mypink}{rgb}{.99,.91,.95}
\definecolor{mycyan}{cmyk}{.3,0,0,0}

\usepackage{adjustbox}
\usepackage{stfloats}

\newcommand\sotaa{\textcolor{red}}
\newcommand\sotab{\textcolor{blue}}
\definecolor{Gray}{rgb}{0.95, 0.95, 0.95}
\newcolumntype{a}{>{\columncolor{Gray}}c}
\usepackage{bm}

\makeatletter
\newcommand*{\new}{\@ifnextchar\bgroup{\new@}{\color{black}}}
\newcommand*{\new@}[1]{{\textcolor{black}{#1}}}
\makeatother

\usepackage{caption}

\begin{document}

\title{PromptSR: Cascade Prompting for Lightweight Image Super-Resolution}

\author{
  Wenyang Liu, Chen Cai, Jianjun Gao, Kejun Wu,~\IEEEmembership{Senior Member,~IEEE}, Yi Wang,~\IEEEmembership{Member,~IEEE} \\ Kim-Hui Yap,~\IEEEmembership{Senior Member,~IEEE} and Lap-Pui Chau,~\IEEEmembership{Fellow,~IEEE}
  \thanks{Wenyang Liu, Chen Cai, Jianjun Gao and Kim-Hui Yap are with School of Electrical and Electronics Engineering, Nanyang Technological University, Singapore.}
  \thanks{Kejun Wu is with the School of Electronic Information and Communications, Huazhong University of Science and Technology, Wuhan 430074, China.}
  \thanks{Yi Wang and Lap-Pui Chau are with the Department of Electrical and Electronic Engineering, The Hong Kong Polytechnic University, Hong Kong.}
  \thanks{Corresponding author: Kejun Wu.}
  }

\maketitle

\begin{abstract}
\new{Although} the lightweight Vision Transformer has significantly advanced image super-resolution (SR), it faces the inherent challenge of a limited receptive field due to the window-based self-attention modeling. 
The quadratic computational complexity relative to window size restricts its ability to use a large window size for expanding the receptive field while maintaining low computational costs. 
To \new{address} this challenge, we propose PromptSR, a novel prompt-empowered lightweight image SR method.
The core component is the proposed \new{cascade prompting block} (CPB), which enhances global information access and local refinement via three cascaded prompting layers: a \new{global anchor prompting layer} (GAPL) and two \new{local prompting layers} (LPLs).
The GAPL leverages downscaled features as anchors to construct low-dimensional \new{anchor prompts} (APs) through cross-scale attention, significantly reducing computational costs. 
These APs, with enhanced global perception, are then used to provide global prompts, efficiently facilitating long-range token connections.
\new{The two LPLs subsequently} combine category-based self-attention and window-based self-attention to refine the representation in a coarse-to-fine manner. 
They leverage attention maps from the GAPL as additional global prompts, enabling them to perceive features globally at different granularities for adaptive local refinement.
In this way, the proposed CPB effectively combines global priors and local details, significantly enlarging the receptive field while maintaining the low computational costs of our PromptSR.
\new{The} experimental results demonstrate the superiority of our method, \new{which outperforms} state-of-the-art lightweight SR methods in quantitative, qualitative, and complexity evaluations. 
Our code will be released at https://github.com/wenyang001/PromptSR.

% for decreasing computational cost and utilize these APs with a global field to for expanding the receptive field.
% summarizes global priors into low-dimensional features as anchors for decreasing computational cost, and harnesses cross-scale attention to build a set of APs with global information for expanding the receptive field. 

\end{abstract}


\section{Introduction} \label{sec:introduction}
% Single image super-resolution (SR) aims to enhance the quality of low-resolution (LR) images by reconstructing them into clean high-resolution (HR) counterparts. Through the application of SR techniques, users can recover intricate details from images captured by hardware-constrained setups, thereby extending their usability without the requirement for costly upgrades. The exploration of efficient and effective SR algorithms, particularly developed for hardware-constrained setups, continues to be a hot topic in computer vision, with a wide range of applications. 

\IEEEPARstart{I}{mage} super-resolution (SR) aims to recover visually pleasing high-resolution (HR) images from a degraded low-resolution (LR) input.
% Widely applied in fields like medical and remote sensing, extracting detailed information from images captured by limited hardware, thus avoiding costly upgrades. 
% Following Dong et al.'s pioneering introduction of a CNN architecture for SR~\cite{dong2015image}, CNN-based methods became prevalent. Subsequent advancements included the adoption of techniques such as residual learning, dense connections, multi-branch networks, and laplacian pyramid structures, significantly improving SR performance.
% Following Dong et al.'s pioneering introduction of a CNN architecture for SR~\cite{dong2015image}, CNN-based methods~\cite{kim2016accurate, huang2017densely, huang2017densely, zhang2018image} have become the mainstream approach. 
% Despite the significant progress made by CNN-based methods, their limited receptive fields pose inherent challenges. 
Since each LR image may correspond to many possible HR counterparts, image SR is a classic, ill-posed, \new{and} challenging problem.
The past decade has witnessed significant progress~\cite{kim2016accurate, huang2017densely, zhang2018image, gao2018image, Ahn_2018_carn, Hui_2019_imdn, Li_2020_lapar} in CNN-based methods for addressing the image SR problem. 
However, these methods struggle to capture long-range dependencies due to the inherent local connectivity of CNNs, which limits their performance.
Recently, the emerging Vision Transformer (ViT) framework~\cite{liang2021swinir, Choi_2022_swinirng,sun2022shufflemixer, STI, srformer, omni_sr, HPI} has shown promising performance in SR by using self-attention to effectively model the long-range \new{relationships} between pixels.
However, this promising performance comes with quadratic computational complexity and memory requirements caused by the self-attention mechanism. 
As a result, these large-scale ViT-based SR models are challenging to deploy on resource-constrained devices in practical applications~\cite{zhang2021edge}, motivating the exploration of lightweight ViT-based SR models (typically, \new{fewer than 1 M} parameters~\cite{omni_sr}).

\begin{figure*}[t]
    \centering    
    \includegraphics[width=0.9\textwidth]{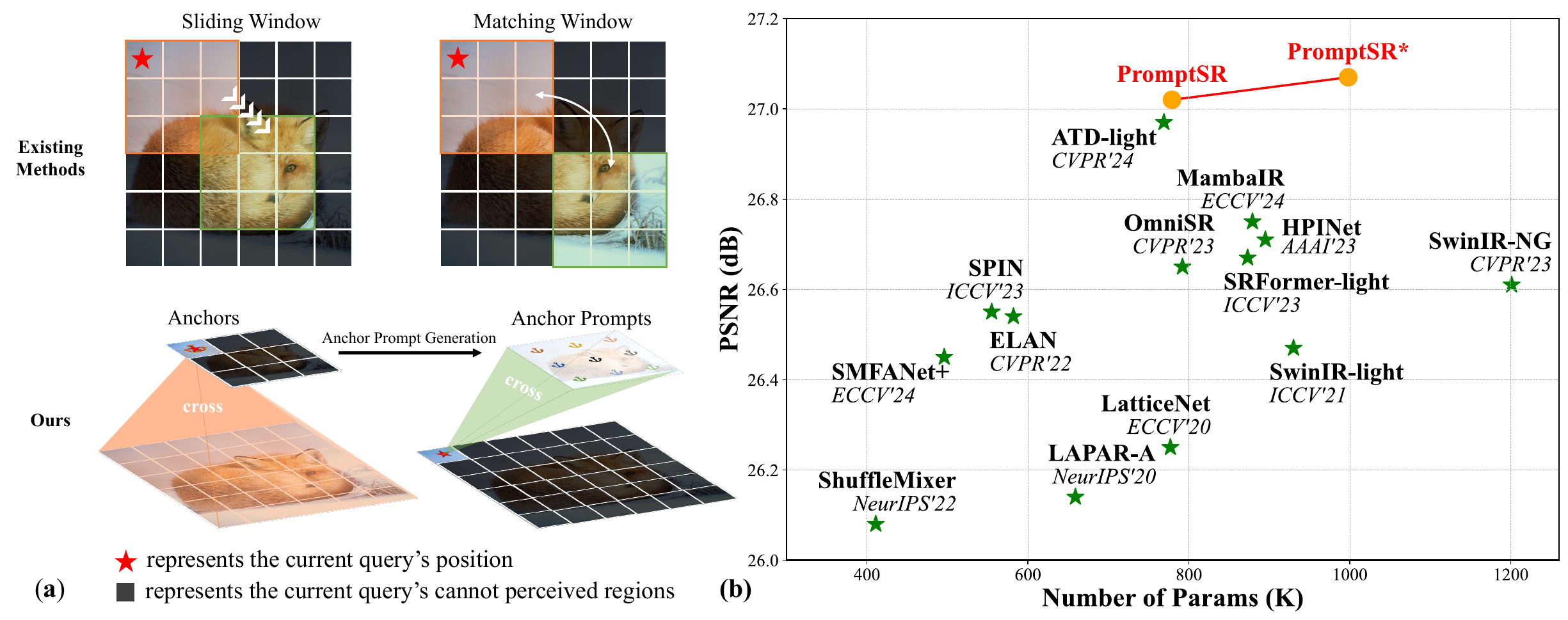}	
    \vspace{-0.05in}I
    \caption{(a) Prevalent window-based attention strategies, shown \textcolor{black}{at} the top row with different colored areas representing windows of adjacent layers, inherently have a limited receptive field at each query position. In contrast, our approach enables global perception \textcolor{black}{via anchor prompts}. Cross: cross-attention. (b) PSNR v.s. Model size for Urban100~\cite{U100} ($\times$4) between our proposed models with varying numbers of \textcolor{black}{cascade prompting blocks} (CPBs) and the latest state-of-the-art lightweight methods. Specifically, we evaluate PromptSR and its enhanced version PromptSR*, to analyze the trade-off between performance and model size.}
    \vspace{-0.2in}
    \label{fig:moti}
\end{figure*}

Recently, window-based attention~\cite{liang2021swinir, Choi_2022_swinirng} has been introduced in Transformers to reduce \new{the} computational complexity and memory requirements for lightweight SR models by limiting attention to fixed-size windows. However, one considerable problem is that the receptive field becomes significantly limited as the window size decreases.
\new{Tokens} within the window, denoted as a red star in Fig.~\ref{fig:moti}(a), to access external information beyond the window for a global receptive field \new{via} two \new{existing} approaches, sliding window-based and matching window-based \new{approaches}. Sliding window-based approaches rely on window-based self-attention with proposed window partitioning and sliding strategies. 
One notable work is SwinIR~\cite{liang2021swinir}, which performs self-attention within fixed-size windows and adopts a sliding window mechanism to facilitate cross-window connections through nearby windows, as shown \new{at} the top-left of Fig.~\ref{fig:moti}(a). 
Although the receptive field can be gradually enlarged as the layer \new{deepens}, effectively capturing dependencies between distant image tokens remains challenging.
Recent SR studies~\cite{HPI, STI} have shifted \new{toward} matching window-based approaches, aiming to select the most suitable window globally based on \new{the basis of} specific matching measurements (e.g., similarity).
% As shown in Fig.~\ref{fig:moti}(a), the image token query within the orange rectangle can freely access corresponding information within the green rectangle through cross-attention mechanisms, enabling the explicit introduction of external information for each local window.
As shown \new{at} the top-right of Fig.~\ref{fig:moti}(a), the query token,  denoted as a red star, within the orange rectangle can freely access the token at the corresponding position within the green rectangle through cross-attention mechanisms, enabling the explicit introduction of external information into the orange local window.
However, these mentioned window-based strategies in lightweight SR still cannot sufficiently exploit global information across the entire feature space, particularly when the window is small (typically 8$\times$8) to reduce computational costs, which inevitably leads to a limited receptive field. 

% A conflict exists between the receptive field and computational cost in the lightweight image SR task.

% This design facilitates the effective integration of global context with local features, resulting in improved SR performance.

Recently, prompting has gained popularity, especially in NLP~\cite{liu2023pre}, and \new{prompting} is increasingly being applied in computer vision tasks~\cite{potlapalli2023promptir}. It aims to introduce additional in-context information to conditionally guide models in performing downstream tasks. Building on this concept, our work extends prompting to lightweight image SR. In this paper, we propose PromptSR, a novel prompting-empowered lightweight image SR method that focuses on enhancing both global and local information perception to address these challenges. 
The core component of the PromptSR is the proposed \new{cascade prompting block} (CPB), which consists of three prompting layers. The first layer, the \new{global anchor prompting layer} (GAPL), introduces \new{anchor prompts} (APs) for global prompting to enhance global information access. The subsequent two layers, the \new{local prompting layers} (LPLs), utilize attention maps from the GAPL as additional prompts for local prompting to progressively refine \new{the} local details.
Specifically, the GAPL capitalizes on the inherent cross-scale similarity property observed in natural images~\cite{anchor}, where basic structures, such as lines, remain consistent across different scaling factors. 
It utilizes downscaled images as \new{anchors}, as shown at the bottom of Fig.~\ref{fig:moti}(a), to generate a set of APs with enhanced global perception.
These APs, with the number matching the size of the local window (as shown at the top of Fig.~\ref{fig:moti}(a)), enable global prompting for each pixel to incorporate out-of-the-window priors through cross-attention, thereby mitigating the limitations of \new{the} local window-based methods while preserving computational efficiency.
The next two LPLs combine window-based self-attention (WSA) and category-based self-attention (CSA) to refine local details in a coarse-to-fine manner.
The WSA is adopted to capture the dependencies within the local image patches, ensuring that fine-grained features are adequately represented. \new{Moreover}, CSA uses attention maps from the GAPL as additional prompts to cluster similar features at different granularities, enabling adaptive local refinement.
Cascading these three prompting layers, the proposed CPB effectively integrates global context with local details, enabling PromptSR to achieve impressive performance, as demonstrated in Fig.~\ref{fig:moti}(b). Our contributions can be summarized as follows:

\begin{enumerate}
    \item We propose PromptSR, a novel lightweight image SR method empowered by prompting. The key insight lies in introducing a set of \new{anchor prompts} (APs) that capitalize on the cross-scale similarity property to capture global priors effectively. By interacting APs with image pixels across the entire space, PromptSR facilitates efficient long-range token interactions, overcoming the constraints imposed by local window partitioning.
    \item We propose the \new{cascade prompting block} (CPB), the fundamental component of our PromptSR. It effectively enlarges the receptive field by cascading three \new{prompt-empowered} layers, enabling feature extraction in a global-to-local and coarse-to-fine manner.
    % : the first layer promotes global information access, while the subsequent layers refine local details in a coarse-to-fine manner, significantly enlarging the receptive field.
    % We exploit the two attentions from the first global anchor prompting layer to guide the next two local prompting layers to achieve local feature refinement in a coarse to fine manner.
    % Unlike existing ViT-based approaches that combine various self-attention mechanisms within a single Transformer layer to enlarge receptive fields,
    % The proposed RHTB incorporates two self-attention layers, each combining two types of self-attention mechanisms, category-based and window-based. This design allow for the refinement of prompted feature representation in a coarse to fine manner.
    \item 
    % Extensive experiments conducted on five public benchmarks validate the superiority of our PromptSR method. It greatly surpasses SOTA lightweight SR methods in quantitative, qualitative, and complexity evaluations. 
    Extensive experiments on five public benchmarks validate the superiority of our PromptSR, demonstrating significant improvements over state-of-the-art lightweight SR methods in quantitative, qualitative, and complexity evaluations. 
    \new{In addition}, our PromptSR (0.78 M parameters) achieves comparable performance to the state-of-the-art large SR model (12 M parameters).
    % Further, our method achieves an excellent model complexity versus performance trade-off.
\end{enumerate}

\section{Related Works}
\subsection{Deep Networks for Image SR} 
% In the past decade, numerous efforts have been made to improve the performance of deep learning techniques in several domains, including super-resolution.
% Kim et al.~\cite{kim2016accurate} utilizes a very deep convolutional network.
% which initially utilized three straightforward convolution layers to convert low-resolution images into high-resolution ones,

In the past decade, numerous efforts have been made to enhance the performance of deep learning techniques in various low-level vision tasks~\cite{liu2023bitstream, liu2023tayi, wu2025end}. Among these \new{methods}, super-resolution (SR) has \new{undergone} significant advancements.
Since the introduction of \new{the} SRCNN~\cite{dong2015image}, a variety of CNN-based networks~\cite{kim2016accurate, huang2017densely, lim2017enhanced, zhang2018image, zhang2020accurate, wang2022ddistill} \new{have demonstrated} advanced state-of-the-art performances through improved architectural designs. Lim~\textit{et al.}~\cite{lim2017enhanced} \new{introduced a} residual block with \new{the} skip connection to enable a deeper network. Zhang~\textit{et al.}~\cite{huang2017densely} further \new{improved} it with a residual dense network (RDN) to connect convolutional blocks. In addition, with the introduction of channel attention, the proposed RCAN~\cite{zhang2018image} extends the network over 400 layers. To obtain comparable reconstruction quality with limited computing resources, several methods \new{have explored} lightweight architectural \new{designs}, such as CARN~\cite{Ahn_2018_carn}, \new{which} leverages a cascading mechanism in a residual network; IMDN~\cite{Hui_2019_imdn}, \new{which} proposes an information distillation block; LAPAR-A~\cite{Li_2020_lapar}, \new{which} constructs a set of predefined filter bases and aims to learn the linear combination coefficients of these filter bases through dictionary learning, and LatticeNet~\cite{Luo_2020_latticenet}, \new{which} proposes the lattice block (LB), a lightweight structure where two butterfly structures are applied to combine two commonly used residual blocks.
Other strategies, \new{such as} large kernel convolution~\cite{luo2016understanding,ding2022scaling}, \new{depthwise} convolution~\cite{hung2019real,sun2022shufflemixer}, self-modulation feature aggregation~\cite{SMFANet}, \new{the} CNN-Transformer hybrid structure~\cite{wang2024camixersr} and \new{the} Mamba SSM architecture~\cite{mambair} are also employed to replace certain computationally expensive components, enabling a more efficient network design.  

\subsection{Lightweight Transformer for Image SR}
With the development of ViT approaches~\cite{dosovitskiy2020image}, the urgent need to apply these models to resource-constrained devices, coupled with concerns about the associated computational costs, has become an issue, motivating the exploration of lightweight Transformers~\cite{liu2024bytenet}. Many attempts~\cite{yang2022lite, huang2022lightvit, sun2022shufflemixer, park2021dynamic, li2023cross} have been made to develop lightweight ViTs. For example, Yang \textit{et al.}~\cite{yang2022lite} proposed LVT, \new{which} incorporates convolution in the self-attention mechanism to enhance the low-level features and reduce computation overhead.
% By replacing the matrix calculation in convolutional layers with transformer layers, MobileViT~\cite{mehta2021mobilevit} achieves better global information capturing.
% EdgeViT~\cite{chen2022edgevit} proposed an information exchange bottleneck to achieve fully spatial interactions, enabling better information communication.
Li \textit{et al.}~\cite{li2023cross} introduced context reasoning into \new{the} Transformer and proposed \new{a} CFIN that can adaptively modify the network weights to selectively incorporate \new{the} desired contextual information. 
Building on these efforts in lightweight Transformers, a variety of techniques have also been developed to enhance the performance of lightweight SR Transformers.
This includes SwinIR~\cite{liang2021swinir} and SwinIR-NG~\cite{liang2021swinir}, \new{which implement} shifted window self-attention; ELAN~\cite{zhang2022efficient}, \new{which uses groupwise multiscale} self-attention; ShuffleMixer~\cite{sun2022shufflemixer}, \new{which leverages} a large \new{depthwise} convolution as \new{a} token mixer; SPIN~\cite{STI}, \new{which proposes} intra-superpixel attention; \new{and} SRFormer~\cite{srformer}, \new{which proposes} permuted self-attention; all aimed at improving the self-attention to achieve better performance.
Furthermore, recent studies \new{such as} OmniSR~\cite{omni_sr} have explored the potential correlations across the omni-axis (i.e., both spatial and channel) to achieve dense information interaction in Transformer layers. HPINet~\cite{HPI} \new{enables} more activated pixels by using window matching and globally mining potential correlations among windows.
Nevertheless, these approaches are mainly based on window-based attention strategies that inevitably suffer from inherent local windows. 
% add some works 
To overcome the limitation of local windows, some works, such as ATD~\cite{ATD1}, introduced dictionary learning in \new{the} Transformer-based approaches, leveraging the auxiliary dictionary to capture global information \new{in the} image tokens. In contrast, other approaches, \new{such as} IPG~\cite{tian2024image}, have shifted to graph neural networks (GNNs), \new{which utilize} the constructed pixel graphs to aggregate global relations.
Our proposed method addresses this limitation by introducing APs for global prompting. Unlike the auxiliary parameters used in ATD~\cite{ATD1}, the APs in our method, with enhanced global perception, are generated by leveraging the inherent cross-scale similarity in natural images, enlarging the receptive field beyond local windows via cross-attention between APs and image tokens.

% ~\cite{yang2022lite, huang2022lightvit, sun2022shufflemixer, Choi_2022_swinirng, HPI}.
% In prior works, it has been proved that combining convolution and Transformers layers can learn both global and local representations. 
% In LVT~\cite{yang2022lite}, the authors incorporate convolution in the self-attention mechanism to enhance the low-level features and reduce computation overhead.
% LightViT~\cite{huang2022lightvit} further proposed an aggregated self-attention mechanism to achieve efficient information aggregation without convolutional computation.

\subsection{Prompt Learning}
Prompt-based learning was initially proposed for transfer learning in NLP~\cite{lester2021power, liu2023pre, li2021prefix}, which involves incorporating additional in-context information to conditionally guide models in performing downstream tasks~\cite{cai2024empowering}. 
Typically, for prompt learning, the learnable prompts are added to input features to direct model predictions and enable more efficient adaptation of parameters. 
Recently, it has been adapted for low-level visual tasks~\cite{potlapalli2023promptir, liu2023unifying, zhou2024seeing, PromptAAAI25, yang2024domain, liu2024dynamic} to effectively guide the models in learning visual content. Potlapalli~\textit{et al.}~\cite{potlapalli2023promptir} developed a generic image restoration model that dynamically restores inputs through interaction with prompts, which act as an adaptive and lightweight module to encode \new{the} degradation context. 
Liu~\textit{et al.}~\cite{liu2023unifying} proposed PromptGIP, a universal framework for general image processing with a visual prompting question-answering paradigm that can handle different image processing tasks.
Similarly, Zhang~\textit{et al.}~\cite{zhang2024docres} proposed DocRes, a visual prompt approach that unifies five document image restoration tasks. Zhou~\textit{et al.}~\cite{zhou2024seeing} further proposed a frequency-prompting image restoration method that uses frequency information to provide prompts.
Nevertheless, few works consider prompt learning for lightweight image SR. 
Here, we propose a novel SR method that harnesses the power of prompting, with the key insight \new{being the introduction of} a set of APs with enhanced global perception to provide global priors for the image SR process.

% using low-dimensional features as anchors to provide global guidance for the image SR process.
% In contrast...
\begin{figure*}[t]
    \centering
\includegraphics[width=\textwidth]{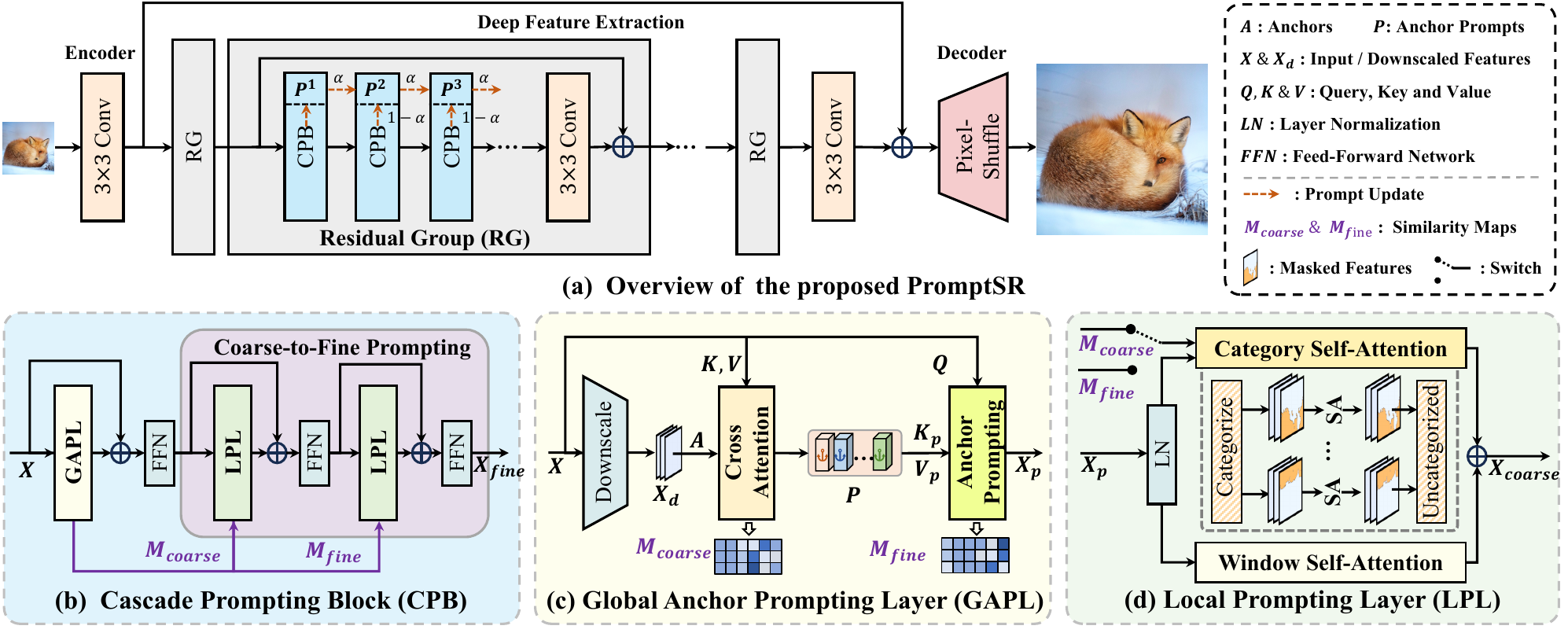}
    \caption{The detailed architecture of (a) our proposed PromptSR, and the structure of (b) the core component of CPB, (c) the GAPL, and (d) the LPL. GAPL leverages the constructed Anchor Prompts $\bm{P}$ to provide global prompts, while LPL utilizes attention maps, i.e., $\bm{M}_{coarse}$ or $\bm{M}_{fine}$ from GAPL as extra prompts. Additionally, Anchor Prompt Update is employed to retain the captured global priors from the previous CPB control by the weighting factor $\alpha$, enhancing consistent feature representations across different CPBs. The cascade of GAPL and LPLs greatly enlarges the receptive field. }
    \label{fig:net} 
    \vspace{-0.2in}
\end{figure*}

\section{Method}
\subsection{Network Architecture}
\textit{\textbf{Overall Structure.}} As shown in Fig.~\ref{fig:net} (a), our proposed PromptSR \new{method} consists of three parts: a shallow feature encoder, a deep feature extraction part, and a pixel-shuffle decoder.
Given the input image $\bm{I}_{LR} \in \mathbb{R}^{H \times W \times 3}$, the shallow feature encoder \new{maps} it to feature embeddings as:

{\small
\begin{equation}
\bm{X}_{0} = \bm{f}_{\operatorname{encoder}}(\bm{I}_{LR}) \in \mathbb{R}^{H \times W \times C} 
\end{equation}
}

\noindent where $C$ denotes the channel number of feature embeddings.
%The shallow feature encoder aims to map the input image $\bm{I}_{LR} \in \mathbb{R}^{H \times W \times 3}$ to feature embeddings $\bm{X}_{0} = \bm{f}_{\operatorname{encoder}}(\bm{I}_{LR}) \in \mathbb{R}^{H \times W \times C}$, where $C$ denote the channel number the embedded features.
$\bm{X_{0}}$ is sent to a series of $K$ \new{residual groups} (RGs) and a $3 \times 3$ convolution to extract deep features $\bm{X_{df}}$ as: 

\vspace{-0.1in}
\begin{small}
\begin{gather}
\bm{X}_{i} = f_{\operatorname{RG_i}}(\bm{X}_{i-1}) \in \mathbb{R}^{H \times W \times C}, \quad \mathbf{i} = 1,2,\cdots, K \\
\bm{X}_{df} = \operatorname{Conv_{3\times3}}(\bm{X}_{K}) \in \mathbb{R}^{H \times W \times C} 
\end{gather}
\end{small}
    
\noindent The RG consists of $N$ \new{cascade prompting blocks} (CPBs) followed by a $3 \times 3$ convolution. Within each CPB, the global APs progressively integrate global priors into output features to enlarge the receptive field. Additionally, the constructed global APs are adaptively updated with APs from the previous block, ensuring the preservation of critical global information across multiple blocks within each RG. Finally, the summation of $\bm{X}_{df}$ and $\bm{X_0}$ is fed into the decoder to reconstruct the high-resolution output with the scale factor $s$: 

\begin{small}
    \begin{equation}
    \bm{I}_{SR} = \bm{f}_{\operatorname{decoder}}(\bm{X}_{df} + \bm{X}_{0}) \in \mathbb{R}^{sH \times sW \times 3}
    \end{equation}
\end{small}

\textit{\textbf{Cascade Prompting Block.}}
% \subsection{Cascade Prompting Block}
The core component of our PromptSR is the proposed CPB, which aims to overcome the inherent limited receptive field constraint imposed by window-based Transformers.
The overall structure is illustrated in Fig.~\ref{fig:net} (b). 
Our CPB differs from \new{the} Transformer blocks used in existing SR methods~\cite{HAT, ttst}, which primarily combine various self-attention mechanisms within a single Transformer layer and stack these layers to implicitly enlarge the receptive field.
Instead, our CPB employs a hierarchical design that cascades a \new{global anchor prompting layer} (GAPL) and two \new{local prompting layers} (LPLs), enabling feature extraction in a global-to-local and coarse-to-fine manner.
This hierarchical design facilitates efficient long-range token interactions, explicitly enabling each token to incorporate out-of-the-window priors and significantly enlarging the receptive field.
More details of the structure of the proposed CPB will be given in the following subsections.

% \subsection{Training Loss}
\textit{\textbf{Training Loss.}}
Following prior works~\cite{liang2021swinir, HAT, HPI, STI}, the standard $\bm{L}_1$ loss between \new{the} SR prediction $\bm{I}_{SR}$ and ground truth HR $\bm{I}_{HR}$ is only used to train our models as \new{follows}:

\begin{small}
\begin{equation}
     \mathcal{L} = ||\bm{I}_{HR} - \bm{I}_{SR}||_1
\end{equation}
\end{small}

%The main contribution of our work is the proposed cascade prompting block (CPB), which is a plug-in module, and architecture agnostic. We present details about the structure as shown in Fig.~\ref{fig:net}(b). 
%We aim to overcome the inherent constraints of window partitioning strategies in Transformers to activate more pixels in our proposed CPB.
%Unlike existing SISR works~\cite{HAT, ttst} that combine various self-attention mechanisms in a single Transformer layer to implicitly expand the receptive field, our proposed CPB employs a different strategy. We stack three prompting layer instantiated with the Transformer structure and utilize both cross-attention and self-attention techniques to enable a hierarchical structure that can explicitly capture from global to local and from coarse to fine features.
% our proposed RHTB stacks more Transformer layers and utilizes both cross-attention and self-attention techniques to build a hierarchical structure from global to local and from coarse to fine. 
% This allows the prompted tokens to adaptively access global priors and hence enhance the global receptive field. More details will be described next. 

\subsection{Global Anchor Prompting Layer}
The objective of the GAPL is to incorporate extensive global priors, overcoming the constraints imposed by local window partitioning while maintaining low computational complexity for lightweight image SR.
Most existing methods utilize effective self-attention mechanisms~\cite{liang2021swinir, Choi_2022_swinirng, HAT} or cross-attention mechanisms~\cite{HPI, chen2021crossvit, HAT}, which are typically window-based. They generate query, key, and value features solely from fixed \textit{local} partitioned windows.
In contrast, we perform cross-attention reasoning with a set of \textit{global} anchor prompts (APs) to capture global priors for effective and efficient image SR. 
Specifically, we first utilize the downscaled input as anchors to generate \new{low-dimensional} global APs and \new{then} utilize the constructed APs to provide global prompts through the cross-attention mechanism. 
By interacting with these APs, we can model long-range dependencies with a \new{computational} overhead of $1/d^2$ compared to self-attention over the entire space, enabling the efficient integration of global priors.
The overall \new{GAPL process} comprises three parts.

% By interacting these APs, it allows us to model long-range dependencies with $1/d^2$ computation overhead, enabling the efficient integration of global priors.

% This layer is constructed using Anchor Prompt Cross-Attention (APCA) to incorporate large global priors while maintaining the lower computational complexity of lightweight image SR.  
% Cross attention mechanism~\cite{HPI, chen2021crossvit, HAT} has been proven to be one of the most effective methods to incorporate or mix extra information in the fields of computer vision and image processing. 
% Inspired by this, in contrast to existing WSA works~\cite{liang2021swinir, Choi_2022_swinirng, HAT} that generate query, key, and value features solely from the input feature, 
% We introduce extra learnable anchor features from the downscaled input feature, which are used to summarize global priors during the training and generate useful anchor prompts for effectively integrating these global priors. 
% The overall process comprises two parts.
% in the $l$-th CPB of an RG
% \textbf{Prompt Construction.} 
\textit{\textbf{Anchor Prompt Generation.}}
% Firstly, the input feature $\bm{X} \in \mathbb{R}^{H \times W \times C}$ is used to generate the query, key, and value features as $\bm{Q, K, V} \in \mathbb{R}^{HW\times C}$ via linear transformation, respectively.
The input feature $\bm{X} \in \mathbb{R}^{H \times W \times C}$ first undergoes linear transformations to generate the query, key, and value features, represented as $\bm{Q, K, V} \in \mathbb{R}^{HW \times C}$.
After that, we downscale $\bm{X}$ to $\bm{X}_{d} \in \mathbb{R}^{H/d \times W/d \times C}$ by a factor $d$ and project $\bm{X}_{d}$ to generate linear embedding as Anchors $\bm{A} \in \mathbb{R}^{HW/d^2 \times C}$ in a similar way as described in~\cite{anchor}.
% Finally, we use the key and the value features to enhance the $\bm{F}$ for constructing anchor prompts $\bm{P} \in \mathbb{R}^{HW/d^2 \times C}$ via cross-attention calculation, formulated as:
% we then use the key and the value features to enhance the $\bm{F}$ for constructing anchor prompts $\bm{P} \in \mathbb{R}^{HW/d^2 \times C}$ via cross-attention calculation, formulated as:
We then use the key $\bm{K}$ and value $\bm{V}$ features from image tokens across the entire space of $\bm{X}$ to enhance Anchors $\bm{A}$ via the cross-attention mechanism. The corresponding outputs form the low-dimensional APs $\bm{P} \in \mathbb{R}^{HW/d^2 \times C}$, resulting in enhanced representations with global perception capabilities.
The process is formulated as follows:
% image tokens from the entire space to query their key and value features to construct low-dimensional Anchor Prompts of $\bm{P} \in \mathbb{R}^{HW/d^2 \times C}$ through the cross-attention mechanism for enhanced global representations, formulated as follows:
% % \vspace{-0.4cm}
% \begin{gather}
%     \mathbf{Q = XW^Q, \quad K = XW^K, \quad V = XW^V, \quad AF=X_dW^{AF}} \\
%     \mathbf{M_{coarse} = \operatorname{SoftMax}((AF)K^T / \sqrt{C}), \quad  AP = M_{coarse}V}
% \end{gather}

\vspace{-0.15in}
\begin{small}
    \begin{gather}
 \bm{Q} = \bm{X}\bm{W}^Q, \  \bm{K} =  \bm{X}\bm{W}^K, \ \bm{V} = \bm{X} \bm{W}^V, \ \bm{A}= \bm{X}_d \bm{W}^{A} \\
\bm{M}_{coarse} =  \bm{A} \bm{K}^T / \sqrt{C}, \ \ 
\bm{P} = \operatorname{SoftMax}( \bm{M}_{coarse}) \cdot  \bm{V}
\end{gather}
\end{small}

\noindent where $\bm{W}^Q, \bm{W}^K, \bm{W}^V, \bm{W}^A \in \mathbb{R}^{C \times C}$ and $\sqrt{C}$ \new{are} the scaled \new{factors} to adjust the range of similarity \new{values}, $\bm{M}_{coarse} \in \mathbb{R}^{HW/d^2 \times HW}$.
\new{Because} query $\bm{A}$ comes from the downscaled image space, which can be regarded as the coarse features of the original images, $\bm{M}_{coarse}$ represents the \textbf{coarse similarity map}, calculated using the dot product between the Anchors $\bm{A}$ and the key features $\bm{K}$.
We then apply a $\operatorname{SoftMax}$ function to transform the similarity map $\bm{M}_{coarse}$ to the attention map, which is further multiplied with the corresponding value $\bm{V}$ features to construct $\bm{P}$.

% Since the query is from the entire space of Anchors $\bm{A}$ instead of local windows, the corresponding outputs $\bm{P}$ can effectively aggregate the long-range dependencies without being constrained within local windows. 
% Since the constructed $P$

%Compared to the input feature $\bm{X}$, the dimension of constructed $\bm{P^l}$ is largely reduced, which enables  $\bm{P^l}$ to adaptively preserve the most important features and filter out some less important local details and even some redundant representation mentioned in~\cite{ttst}. We believe these preserved features in $\bm{P^l}$ during each Residual Group (RG) may have some common representation and contain consistent external priors. To enhance these consistent representations, we introduce an adaptive prompt update process in each RG to let $\bm{P}$ be updated from the last CPB, formulated as follows: 

% \textbf{Anchor Prompt Update.} 
% Compared to the input feature $\bm{X}$, the dimension of the constructed $\bm{P^l}$ is significantly reduced. This reduction enables $\bm{P}^l$ to adaptively preserve the most important features while filtering out less important local details and even some redundant representations that are mentioned in~\cite{ttst}.

\textit{\textbf{Anchor Prompt Update.}} Compared \new{with the values of} $\bm{Q, K,}$ and $\bm{V}$ from the input feature $\bm{X}$, the dimension of the constructed $\bm{P}$ is significantly reduced by $1/d^2$. 
This reduction allows $\bm{P}$ to adaptively preserve the most crucial features while filtering out less important features from $\bm{X}$, such as \new{the} local details and redundant representations.
By doing so, we believe that the preserved information in $\bm{P}$ across different CPBs can exhibit similar feature representations, capturing consistent global priors. 
Therefore, we propose an adaptive anchor prompt update across different CPBs to enhance these consistent feature representations, allowing $\bm{P}$ to be adaptively updated within the last CPB.
However, existing studies~\cite{dong2015image, liang2021swinir} \new{have} indicated that as \new{the depth of deep learning models increases}, the features extracted exhibit varying characteristics. For \new{example}, in shallow layers, the network primarily focuses on basic features such as edges and textures; as the depth increases, the model gradually learns more complex features such as shapes and structures. 
Thus, our proposed update process is restricted within each RG while avoiding updates across different RGs to preserve the diversity of captured features in $\bm{P}$.
% To enhance these consistent representations within RGs while preserving the diversity across different RGs, we introduce an adaptive prompt update process within each RG, allowing $\bm{P}$ to be updated from the last CPB. 
This process is formulated as follows:

\begin{small}
    \begin{equation}
        \bm{P}^i = \alpha \bm{P}^{i-1} + (1-\alpha)\bm{P}^{i}
    \end{equation}
\end{small}

\noindent where $\bm{P}^{i-1}$ and $\bm{P}^i$ represent constructed APs from the CPB of $i-1$ and $i$ within single RG, respectively, and $\alpha$ represents a small constant that controls the weight assigned to the last constructed APs of $i-1$.

\begin{figure}[t]
    \centering
\includegraphics[width=0.47\textwidth]{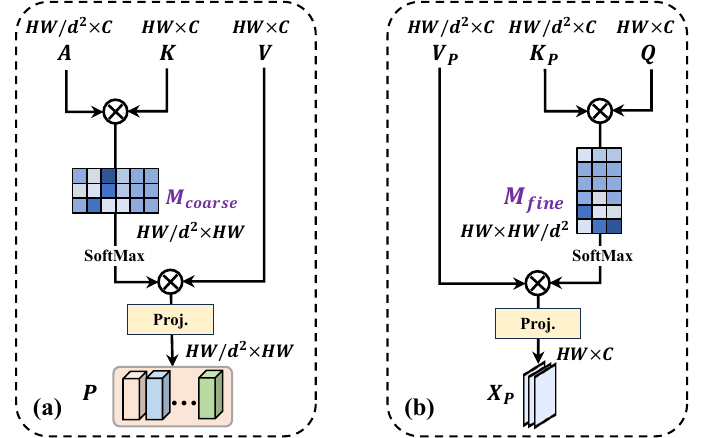}
    \caption{The extraction of similarity maps: (a) coarse similarity map (b) fine similarity map. The Proj. operation is implemented using a convolutional layer for output projection to enhance the representational capability. }
    \label{fig:attn} 
    \vspace{-0.2in}
\end{figure}

% \vspace{-0.4cm}
% \begin{equation}
%     \bm{P}^l = \alpha \bm{P}^{l-1} + (1-\alpha)\bm{P}^{l}
% \label{eq:update}
% \end{equation}

% \noindent where $\bm{P}^{l-1}$ and $\bm{P}^l$ represent the generated anchor prompts from the CPB of $l-1$ and $l$, respectively, and $\alpha$ represents a small constant to control the weight of the last generated anchor prompts.
 
% In this way, 
% By using anchor tokens $A$ from the downscaled image to query key tokens $K$ from the entire original image, the receptive field is fully expanded

\textit{\textbf{Anchor Prompting.}} 
% To enable the constructed anchor prompts $\bm{P}$ to interact with the input feature $\bm{X}$, we use the anchor prompts to generate the new key and value features as $\bm{K}_{p}, \bm{V}_{p} \in \mathbb{R}^{HW/d^2 \times C}$ features via linear transformation. After that, the generated $\bm{K}_{p}$ and $\bm{V_{p}}$ will be utilized to prompt the value features $\bm{V}$ from $\bm{X}$ to capture global priors via cross-attention calculation, 
To enable the constructed APs $\bm{P}$ to provide global prompts for the input feature $\bm{X}$, we generate new key and value features, $\bm{K}_{p},\bm{V}_{p} \in \mathbb{R}^{HW/d^2 \times C}$, from $\bm{P}$ through \new{a} linear transformation.
These features, $\bm{K}_{p}$ and $\bm{V}_{p}$, are then used to prompt the value features $\bm{Q}$ from the input feature $\bm{X}$ via \new{a} cross-attention calculation, formulated as:

% \vspace{-4mm}
% \begin{gather}
%    \mathbf{K_{ap} = (AP)W^{K_{ap}}, \quad V_{ap} = (AP) W^{V_{ap}}} \\
%   \mathbf{M_{fine} = \operatorname{SoftMax}((Q(K^T_{ap} + AF^T)/ \sqrt{C}) \quad X_p = M_{fine}V_{ap}}
% \end{gather}

\vspace{-0.15in}
\begin{small}
  \begin{gather}
    \bm{K}_{p} = \bm{P} \bm{W}^{K_{p}}, \ \bm{V}_{p} =  \bm{P} \bm{W}^{V_{p}} \\
    \bm{M}_{fine} =  \bm{Q}( \bm{K}_{p} +  \bm{A})^T / \sqrt{C}, \ \
    \bm{X}_p = \operatorname{SoftMax}(\bm{M}_{fine}) \cdot  \bm{V}_{p}
\end{gather}  
\end{small}

\noindent \new{where} $\bm{W}^{K_{p}}, \bm{W}^{V_{p}} \in \mathbb{R}^{C \times C}$, and $\bm{M}_{fine} \in \mathbb{R}^{HW \times HW/d^2}$. 
Since the query $Q$ is from the original image space, it can be regarded as the fine features of the original images. $\bm{M}_{fine}$ represents the \textbf{fine similarity map}, \new{which is} calculated using the dot product between \new{the} query features $\bm{Q}$ and both the Anchors $\bm{A}$ and the key $\bm{K}_{p}$ features.
We then apply a $\operatorname{SoftMax}$ function to transform the similarity map $\bm{M}_{fine}$ to the attention map, which is further multiplied with the corresponding value $\bm{V}_p$ features to generate the prompted features $\bm{X}_p$.
By doing so, the prompted features $\bm{X}_p$ achieve enhanced global representations through APs, enabling a global receptive field and overcoming the constraints imposed by local window partitioning.

To elaborate on the process of generating the coarse and fine similarity maps, we further provide a diagram in Fig~\ref{fig:attn}.  
Furthermore, after the GAPL processing, we incorporate commonly used LayerNorm and feed-forward network (FFN) layers to enhance the feature expression of $\bm{X}_{p}$, formulated as:

% \vspace{-0.4cm}
% \begin{gather}
%   \bm{X}_p = \bm{X}_p + \bm{X} \\
%     \bm{X}_{p} = \operatorname{FFN}(\operatorname{LN}(\bm{X}_p)) + \bm{X}_p
% \label{RTB}
% \end{gather}

% \vspace{-0.15in}
% \begin{small}
%     \begin{equation}
%      \bm{X}_p = \bm{X}_p + \bm{X}, \ \bm{X}_{p} = \operatorname{FFN}(\operatorname{LN}(\bm{X}_p)) + \bm{X}_p
% \end{equation}
% \end{small}

\vspace{-0.15in}
\begin{small}
\begin{gather}
     \bm{X}_p = \bm{X}_p + \bm{X} \\ \bm{X}_{p} = \operatorname{FFN}(\operatorname{LN}(\bm{X}_p)) + \bm{X}_p
\end{gather}
\end{small}

\subsection{Coarse-to-Fine Local Prompting Layers}
As Fig.~\ref{fig:net} (b) shows, our \new{coarse-to-fine prompting} module consists of two LPLs with an identical structure, designed to enhance local details by using extra global priors captured in the last layer to achieve improved reconstruction results.
Each LPL preserves the widely utilized \new{window-based self-attention} (WSA) mechanism~\cite{liang2021swinir}, which operates within fixed-shaped windows to model local image tokens effectively.
We further overcome window-based constraints by integrating \new{category-based self-attention} (CSA)~\cite{ATD1} in LPLs, as shown in Fig.~\ref{fig:net} (d). 
This integration allows our LPLs to leverage similarity maps from the GAPL layer as extra global prompts to guide the feature refinement effectively from a global perspective. 
Specifically, $\bm{M}_{coarse}$ and $\bm{M}_{fine}$ are used successively to form \new{irregularly} shaped windows containing similar/related image tokens through the $\operatorname{Categorization}$ process, enabling the global perception of external features beyond the fixed-shaped windows.
Following this, the self-attention technique is applied to each \new{irregularly} shaped window to refine these image features. 
Finally, these refined features are combined to produce the final output through the $\operatorname{Uncategorization}$ process. 
The overall process mainly comprises two parts.

% To achieve this, we design a Coarse-to-Fine Prompting strategy by integrating Category-based Self-Attention (CSA)~\cite{ATD1} into the proposed LPL.

% Moreover, to integrate global priors for pixel categorization and comprise global semantic information, we employ fine-prompting from the previous GAPL to enhance the recent Category-based Self-Attention (CSA)~\cite{ATD1}. This strategy enriches the global receptive field and uses a similarity map generated from an irregularly shaped window for adaptive feature partitioning. 

\textit{\textbf{Coarse Prompting.}}
Given the anchor-prompted feature $\bm{X}_p$, we first use the calculated coarse similarity map $\bm{M}_{coarse}$ to classify each image token of $\bm{X}_p$ into different categories $\bm{\theta}^1, \bm{\theta}^2, \cdots,\bm{\theta}^M$ \new{on the basis of} $\bm{\theta}^{i} = \{\bm{x}_j|\mathop{\arg\max}_k(\bm{M}_{coarse}^{kj} = i)\}$.  
The image token $\bm{x}_j$ is thus classified into the category $\bm{\theta}^i$ if it achieves the highest similarity score.
In this way, image tokens with the same category can be perceived as being grouped into an irregularly shaped window, where the self-attention technique is further applied to refine the details to \new{obtain} $\bm{X}_{coarse}$. 
The process can be formulated as \new{follows}:

% Given the anchor prompted feature $\bm{X}_p$, we first leverage the calculated coarse similarity map $\bm{M}_{coarse}$ to classify each pixel token of $\bm{X}_p$ into different categories $\bm{\theta}^1, \bm{\theta}^2, \cdots,\bm{\theta}^M$ based on $\bm{\theta}^{i} = \{\bm{x}_j|\mathop{\arg\max}_k(\bm{M}_{coarse}^{kj} = i)\}$. The pixel token $\bm{x}_j$ will be classified into the category $\bm{\theta}_i$ if the similarity score achieves the highest score.
% In this way, pixel tokens with the same category can be perceived as being grouped into an irregularly shaped window, where the self-attention technique is further applied to refine the details to get $\bm{X}_{coarse}$. 
% After that, we further leverage the calculated fine similarity map $\bm{M}_{fine}$ to classify each pixel token of $\bm{X}_{coarse}$ into different categories based on $\bm{\theta}^{i} = \{\bm{x}_j|\mathop{\arg\max}_k(\bm{M}_{coarse}^{jk} = i)\}$. The operation is almost the same as coarse prompting except that the highest score is obtained along different dimensions. 
% The overall process can be formulated as:  
\vspace{-0.15in}
\begin{small}
    \begin{gather}
\bm{X}_{coarse} = \operatorname{WSA}(\bm{X}_{p}) + \operatorname{CSA}(\bm{X}_{p}, \bm{M}_{coarse}) + \bm{X}_{p} \\
\bm{X}_{coarse} = \operatorname{FFN}(\operatorname{LN}(\bm{X}_{coarse})) + \bm{X}_{coarse}
\end{gather}
\end{small}

\textit{\textbf{Fine Prompting.}} After that, we leverage the calculated fine similarity map $\bm{M}_{fine}$ to classify each pixel token of $\bm{X}_{coarse}$ into different categories \new{on the basis of} $\bm{\theta}^{i} = \{\bm{x}_j|\mathop{\arg\max}_k(\bm{M}_{coarse}^{jk} = i)\}$. 
The operation is almost the same as the coarse prompting, except that the highest score is obtained along different dimensions. 
The process can be formulated as \new{follows}:

\vspace{-0.15in}
\begin{small}
\begin{gather}
\bm{X}_{fine} = \operatorname{WSA}(\bm{X}_{coarse}) + \operatorname{CSA}(\bm{X}_{coarse}, \bm{M}_{fine}) + \bm{X}_{coarse} \\
\bm{X}_{fine} = \operatorname{FFN}(\operatorname{LN}(\bm{X}_{fine})) + \bm{X}_{fine}
\label{RTB}
\end{gather}
\end{small}

\noindent By utilizing the calculated similarity maps $\bm{M}_{coarse}$ and $\bm{M}_{fine}$ in LPLs, we can adaptively refine the prompted feature $\bm{X}_p$ in a coarse to fine manner.

\begin{table*}[t]
% \captionsetup{font={small}}
% \scriptsize
% \footnotesize
% \small
% \setlength{\tabcolsep}{2.8pt}
\setlength{\tabcolsep}{1.7mm}
\vspace{-0.2in}
\caption{Quantitative comparison (PSNR/SSIM) with state-of-the-art methods on lightweight SR task. The best and second best results are colored with \sotaa{red} and \sotab{blue}.}
\vspace{-2mm}
\label{tab:results1}
  % \centering
  % \begin{tabular}{l|l}
  \begin{center}
    % \begin{adjustbox}{width=1\textwidth}
\resizebox{0.85\textwidth}{!}{
  \begin{tabular}{l|c|c|cc|cc|cc|cc|cc}
    % \begin{tabular}{lcccccccccccc}
    \bottomrule

    % \cmidrule(r){1-4}
    \multirow{2}{*}{\textbf{Method}} &
    \multirow{2}{*}{\textbf{Scale}} & \multirow{2}{*}{\textbf{\#Params.}} & \multicolumn{2}{c|}{\textbf{Set5}} & \multicolumn{2}{c|}{\textbf{Set14}} & \multicolumn{2}{c|}{\textbf{BSD100}} & \multicolumn{2}{c|}{\textbf{Urban100}} & \multicolumn{2}{c}{\textbf{Manga109}} \\   \cline{4-13}
    % $\times$2 %%%%%%%%%%%%%%%%%%%%%%%%%%%%%%%%%%%%%%%%%%%%%%%%%%%
    & & & PSNR & SSIM & PSNR & SSIM & PSNR & SSIM & PSNR & SSIM & PSNR & SSIM   \\  
   % \midrule
   \hline
   
    IDN~\cite{hui2018fast}   & $\times$2 & 553K   & 37.83 & 0.9600 & 33.30 & 0.9148 & 32.08 & 0.8985 & 31.27 & 0.9196 & 38.01 & 0.9749 \\
    CARN~\cite{Ahn_2018_carn}    &  $\times$2 & 1,592K & 37.76 & 0.9590 & 33.52 & 0.9166 & 32.09 & 0.8978 & 31.92 & 0.9256 & 38.36 & 0.9765 \\ 
    IMDN~\cite{Hui_2019_imdn}    & $\times$2 & 694K   & 38.00 & 0.9605 & 33.63 & 0.9177 & 32.19 & 0.8996 & 32.17 & 0.9283 & 38.88 & 0.9774 \\
    % MADNet~\cite{Hui_2019_imdn}   & $\times$2 & 878K   & 37.85 & 0.9600 & 33.38 & 0.9161 & 32.04 & 0.8979 & 31.62 & 0.9233 & - & -\\
    LAPAR-A~\cite{Li_2020_lapar}  & $\times$2 & 548K   & 38.01 & 0.9605 & 33.62 & 0.9183 & 32.19 & 0.8999 & 32.10 & 0.9283 & 38.67 & 0.9772 \\
    LatticeNet~\cite{Luo_2020_latticenet} & $\times$2 & 756K   & 38.15 & 0.9610 & 33.78 & 0.9193 & 32.25 & 0.9005 & 32.43 & 0.9302 & -     & -      \\
    SwinIR-light~\cite{liang2021swinir} & $\times$2 & 910K   & 38.14 & 0.9611 & 33.86 & 0.9206 & 32.31 & 0.9012 & 32.76 & 0.9340 & 39.12 & 0.9783 \\
    ShuffleMixer~\cite{sun2022shufflemixer} & $\times$2 & 394K & 38.01 & 0.9606 & 33.63 & 0.9180 & 32.17 & 0.8995 & 31.89 & 0.9257 & 38.83 & 0.9774 \\
    ELAN~\cite{zhang2022efficient}   & $\times$2 & 582K   & 38.17 & 0.9611 & 33.94 & 0.9207 & 32.30 & 0.9012 & 32.76 & 0.9340 & 39.11 & 0.9782 \\
    SwinIR-NG~\cite{Choi_2022_swinirng}  & $\times$2 & 1181K  & 38.17 & 0.9612 & 33.94 & 0.9205 & 32.31 & 0.9013 & 32.78 & 0.9340 & 39.20 & 0.9781 \\
    SPIN~\cite{STI}  & $\times$2 & 497K & 38.20 & 0.9615 & 33.90 & 0.9215 & 32.31 & 0.9015 & 32.79 & 0.9340 & 39.18 & 0.9784 \\
    SRFormer-light~\cite{srformer} & $\times$2 & 853K & 38.23 & 0.9613 & 33.94 & 0.9209 & 32.36 & 0.9019 & 32.91 & 0.9353 & 39.28 & 0.9785\\
    OmniSR~\cite{omni_sr}   & $\times$2 & 772K   & 38.22 & 0.9613 & 33.98 &  0.9210 & 32.36 & 0.9020 & 33.05 & 0.9363 & 39.28 & 0.9784 \\

    HPINet~\cite{HPI} &  $\times$2  & 783K  & 38.12 & 0.9605 & 33.94 &  0.9209 & 32.31 & 0.9013 & 32.85 & 0.9346 & 39.08 &  0.9772 \\

    MambaIR~\cite{mambair} & $\times$2 & 859K & 38.16 & 0.9610 & 34.00 & 0.9212 & 32.34  & 0.9017 & 32.92 & 0.9356 & 39.31 & 0.9779 \\
    SMFANet$+$~\cite{SMFANet} & $\times$2 & 480K  & 38.19 & 0.9611 & 33.92 & 0.9207 & 32.32 & 0.9015 & 32.70 & 0.9331 & 39.46 & 0.9787 \\
    % IPG-Tiny~\cite{tian2024image}  & $\times$2 & 872K & 38.27  & \sotab{0.9616} & \sotaa{34.24} & \sotaa{0.9236} & 32.35  & 0.9018  & 33.04 & 0.9359 & 39.31 & 0.9786   \\ 
    CAMixerSR~\cite{wang2024camixersr} &  $\times$2 & 746K  & 38.23 & 0.9613 & 34.00 & 0.9214 & 32.34 & 0.9016 & 32.95 & 0.9348 & 39.32 & 0.9781 \\

    ATD-light~\cite{ATD1}    &  $\times$2 & 753K   & \sotab{38.28} & \sotab{0.9616} & \sotab{34.11} & 0.9217 & \sotaa{32.39} & \sotaa{0.9023} & 33.27 & 0.9376 & 39.51 & 0.9789 \\

    % \rowcolor{Gray}
    % \textbf{HCSA} (Ours)  & $\times$2 & 731K   & \sotaa{38.29} & \sotaa{0.9616} & \sotaa{34.12} & \sotaa{0.9233} & \sotaa{32.39} & \sotaa{0.9022} & \sotaa{33.44} & \sotaa{0.9391} & \sotaa{39.51} & \sotaa{0.9789} \\

    \rowcolor{Gray}
\textbf{PromptSR (Ours)} & $\times$2 & 764K   & \sotaa{38.30} & \sotaa{0.9617} & 34.10 & \sotab{0.9221} & \sotab{32.37} & \sotab{0.9022} & \sotab{33.39} & \sotab{0.9390} & \sotab{39.56} & \sotab{0.9790} \\
    \rowcolor{Gray}
\textbf{PromptSR$^\star$ (Ours)} & $\times$2 & 982K   & \sotaa{38.30} & \sotaa{0.9617} & \sotaa{34.13} & \sotaa{0.9231} & \sotaa{32.39} & \sotaa{0.9023} & \sotaa{33.53} & \sotaa{0.9396} & \sotaa{39.60} & \sotaa{0.9791} \\

    % $\times$3 %%%%%%%%%%%%%%%%%%%%%%%%%%%%%%%%%%%%%%%%%%%%%%%%%%%
    % \midrule
    % \hline \hline
    \hline 
    IDN~\cite{hui2018fast}  & $\times$3 & 553K  & 34.11 & 0.9253 & 29.99 & 0.8354 & 28.95 & 0.8013 & 27.42 & 0.8359 & 32.71 & 0.9381 \\
    CARN~\cite{Ahn_2018_carn} & $\times$3 & 1,592K & 34.29 & 0.9255 & 30.29 & 0.8407 & 29.06 & 0.8034 & 28.06 & 0.8493 & 33.50 & 0.9440 \\
    IMDN~\cite{Hui_2019_imdn} & $\times$3 & 703K   & 34.36 & 0.9270 & 30.32 & 0.8417 & 29.09 & 0.8046 & 28.17 & 0.8519 & 33.61 & 0.9445 \\
    % MADNet~\cite{Hui_2019_imdn}   & $\times$3 & 930K   & 34.16 & 0.9253 & 30.21 & 0.8398 & 28.98 & 0.8023 & 27.77 & 0.8439 & - & -\\
    LAPAR-A~\cite{Li_2020_lapar} & $\times$3 & 544K   & 34.36 & 0.9267 & 30.34 & 0.8421 & 29.11 & 0.8054 & 28.15 & 0.8523 & 33.51 & 0.9441 \\
    LatticeNet~\cite{Luo_2020_latticenet} & $\times$3 & 765K   & 34.53 & 0.9281 & 30.39 & 0.8424 & 29.15 & 0.8059 & 28.33 & 0.8538 & -     & -      \\
    SwinIR-light~\cite{liang2021swinir} & $\times$3 & 918K   & 34.62 & 0.9289 & 30.54 & 0.8463 & 29.20 & 0.8082 & 28.66 & 0.8624 & 33.98 & 0.9478 \\
    ShuffleMixer~\cite{sun2022shufflemixer} & $\times$3 & 415K & 34.40 & 0.9272 & 30.37 & 0.8423 & 29.12 & 0.8051 & 28.08 & 0.8498 & 33.69 & 0.9448 \\
    ELAN~\cite{zhang2022efficient} & $\times$3 & 590K   & 34.61 & 0.9288 & 30.55 & 0.8463 & 29.21 & 0.8081 & 28.69 & 0.8624 & 34.00 & 0.9478 \\
    SwinIR-NG~\cite{Choi_2022_swinirng} & $\times$3 & 1190K  & 34.64 & 0.9293 & 30.58 & 0.8471 & 29.24 & 0.8090 & 28.75 & 0.8639 & 34.22 & 0.9488 \\
    SPIN~\cite{STI} &  $\times$3 & 569K & 34.65 & 0.9293 & 30.57 & 0.8464 & 29.23 & 0.8089 & 28.71 & 0.8627 & 34.24 & 0.9489 \\
    SRFormer-light~\cite{srformer} & $\times$3 & 861K & 34.67 & 0.9296 & 30.57 & 0.8469 & 29.26 & 0.8099 & 28.81 & 0.8655 & 34.19 & 0.9489\\
    OmniSR~\cite{omni_sr}  & $\times$3 & 780K   & 34.70 & 0.9294 & 30.57 & 0.8469 & 29.28 & 0.8094 & 28.84 & 0.8656 & 34.22 & 0.9487 \\
    HPINet~\cite{HPI} & $\times$3 & 924K  &  34.70 & 0.9289 & 30.63 & 0.8480 & 29.26 & 0.8104 & 28.93 & 0.8675 & 34.21 &  0.9488 \\
   MambaIR~\cite{mambair} & $\times$3 & 867K &  34.72 & 0.9296 & 30.63 & 0.8475 & 29.29 & 0.8099 & 29.00 & 0.8689 & 34.39 & 0.9495 \\
   SMFANet$+$~\cite{SMFANet} & $\times$3 & 487K & 34.63 & 0.9285 & 30.52 & 0.8456 & 29.23 & 0.8084 & 28.59 & 0.8594 & 34.17 & 0.9478 \\
% IPG-Tiny~\cite{tian2024image} & $\times$3 & 878K & 34.64 & 0.9292 & 30.61 &    0.8470 & 29.26 & 0.8097 & 28.93  & 0.8666 & 34.30 & 0.9493 \\
    ATD-light~\cite{ATD1}   & $\times$3 & 760K   & 34.70 & \sotab{0.9300} & \sotab{30.68} & 0.8485 & \sotaa{29.32} & \sotaa{0.8109} & 29.16 & 0.8710 & 34.60 & 0.9505 \\
    % \rowcolor{Gray}
    
    % \textbf{HCSA} (Ours) & $\times$3 & 732K   & \sotaa{34.75} & \sotaa{0.9299} & \sotaa{30.71} & \sotaa{0.8491} & \sotaa{29.31} & \sotaa{0.8106} & \sotaa{29.29} & \sotaa{0.8728} & \sotaa{34.60} & \sotaa{0.9507} \\
    \rowcolor{Gray}
    \textbf{PromptSR (Ours)} & $\times$3 & 770K   & \sotab{34.73} & \sotab{0.9300} & \sotab{30.68} & \sotab{0.8486} & \sotab{29.30} & \sotab{0.8105} & \sotab{29.26} & \sotab{0.8725} & \sotab{34.63} & \sotab{0.9507} \\
    \rowcolor{Gray}
\textbf{PromptSR$^\star$ (Ours)} & $\times$3 & 989K   & \sotaa{34.78} & \sotaa{0.9301} & \sotaa{30.74} & \sotaa{0.8493} & \sotaa{29.32} & \sotaa{0.8109} & \sotaa{29.31} & \sotaa{0.8735} & \sotaa{34.71} & \sotaa{0.9511} \\

    % $\times$4 %%%%%%%%%%%%%%%%%%%%%%%%%%%%%%%%%%%%%%%%%%%%%%%%%%%
    % \midrule
    \hline 
    IDN~\cite{hui2018fast}  & $\times$4 & 553K  & 31.82 & 0.8903 & 28.25 & 0.7730 & 27.41 & 0.7297 & 25.41 & 0.7632 & 29.41 & 0.8942 \\
    CARN~\cite{Ahn_2018_carn}  & $\times$4 & 1,592K & 32.13 & 0.8937 & 28.60 & 0.7806 & 27.58 & 0.7349 & 26.07 & 0.7837 & 30.47 & 0.9084 \\
    IMDN~\cite{Hui_2019_imdn} & $\times$4 & 715K   & 32.21 & 0.8948 & 28.58 & 0.7811 & 27.56 & 0.7353 & 26.04 & 0.7838 & 30.45 & 0.9075 \\
    % MADNet~\cite{Hui_2019_imdn}   & $\times$4 & 1002K   & 31.95 & 0.8917 & 28.44 & 0.7780 & 27.47 & 0.7327 & 25.76 & 0.7746 & - & -\\
    LAPAR-A~\cite{Li_2020_lapar} & $\times$4 & 659K   & 32.15 & 0.8944 & 28.61 & 0.7818 & 27.61 & 0.7366 & 26.14 & 0.7871 & 30.42 & 0.9074 \\
    LatticeNet~\cite{Luo_2020_latticenet} & $\times$4 & 777K   & 32.30 & 0.8962 & 28.68 & 0.7830 & 27.62 & 0.7367 & 26.25 & 0.7873 & -     & -      \\
    SwinIR-light~\cite{liang2021swinir} & $\times$4 & 930K   & 32.44 & 0.8976 & 28.77 & 0.7858 & 27.69 & 0.7406 & 26.47 & 0.7980 & 30.92 & 0.9151 \\
    ShuffleMixer~\cite{sun2022shufflemixer} & $\times$4 & 411K & 32.21 & 0.8953 & 28.66 & 0.7827 & 27.61 & 0.7366 & 26.08 & 0.7835 & 30.65 & 0.9093 \\
    ELAN~\cite{zhang2022efficient} & $\times$4 & 582K   & 32.43 & 0.8975 & 28.78 & 0.7858 & 27.69 & 0.7406 & 26.54 & 0.7982 & 30.92 & 0.9150 \\
    SwinIR-NG~\cite{Choi_2022_swinirng}  & $\times$4 & 1201K  & 32.44 & 0.8980 & 28.83 & 0.7870 & 27.73 & 0.7418 & 26.61 & 0.8010 & 31.09 & 0.9161 \\
    SPIN~\cite{STI} & $\times$4 & 555K & 32.48 & 0.8983 & 28.80 & 0.7862 & 27.70 & 0.7415 & 26.55 & 0.7998 & 30.98 & 0.9156 \\
    SRFormer-light~\cite{srformer} & $\times$4 & 873K & 32.51 & 0.8988 & 28.82 & 0.7872 & 27.73 & 0.7422 & 26.67 & 0.8032 & 31.17 & 0.9165 \\
    OmniSR~\cite{omni_sr} & $\times$4 & 792K   & 32.49 & 0.8988 & 28.78 & 0.7859 & 27.71 & 0.7415 & 26.65 & 0.8018 & 31.02 & 0.9151 \\
    HPINet~\cite{HPI}& $\times$4 & 895K   &   32.60 & 0.8986 & 28.87 & 0.7874 & 27.73 & 0.7419 & 26.71 & 0.8043 & 31.19 & 0.9162 \\
     MambaIR~\cite{mambair} & $\times$4 & 879K  & 32.51 &  0.8993 & 28.85 & 0.7876 & 27.75 & 0.7423 & 26.75 & 0.8051 & 31.26 & 0.9175 \\
    SMFANet$+$~\cite{SMFANet} & $\times$4 & 496K & 32.43 & 0.8979 & 28.77 & 0.7849 & 27.70 & 0.7400 & 26.45 & 0.7943 & 31.06 & 0.9138 \\
    % IPG-Tiny~\cite{tian2024image} &  $\times$4 & 887K & 32.51 & 0.8987 & 28.85 &   0.7873 & 27.73  & 0.7418 & 26.78 & 0.8050 & 31.22 & 0.9176        \\

 CAMixerSR~\cite{wang2024camixersr} & $\times$4 & 765K & 32.51 & 0.8988 & 28.82 & 0.7870 & 27.72 & 0.7416 & 26.63 & 0.8012 & 31.18 & 0.9166 \\
    
    ATD-light~\cite{ATD1}   & $\times$4 & 769K   & \sotab{32.62} & 0.8997 & 28.87 & 0.7884 & \sotab{27.77} & \sotab{0.7439} & 26.97 & 0.8107 & 31.47 & \sotab{0.9198} \\

     % \cmidrule{2-15}
    % \rowcolor{Gray}
    %  \textbf{HCSA} (Ours) & $\times$4 & 748K   & 32.56 & \sotaa{0.8993} & \sotaa{28.91} & \sotaa{0.7889} & \sotaa{27.78} & \sotaa{0.7435} & \sotaa{27.03} & \sotaa{0.8122} & \sotaa{31.50} & \sotaa{0.9200} \\
% \cdashline{1-13}
    
    \rowcolor{Gray}
     \textbf{PromptSR (Ours)} & $\times$4 & 779K   & 32.61 & \sotab{0.8999} & \sotab{28.94} & \sotab{0.7892} & \sotab{27.77} & 0.7435 & \sotab{27.02} & \sotab{0.8116} & \sotab{31.50} & \sotab{0.9198} \\
    \rowcolor{Gray}
     \textbf{PromptSR$^\star$ (Ours)} & $\times$4 & 998K   & \sotaa{32.65} & \sotaa{0.9001} & \sotaa{28.95} & \sotaa{0.7895} & \sotaa{27.79} & \sotaa{0.7441} & \sotaa{27.07} & \sotaa{0.8134} & \sotaa{31.57} & \sotaa{0.9205} \\
\toprule
  \end{tabular}
  % \end{adjustbox}
  }
  \end{center}
  \vspace{-0.15in}
\end{table*}

\begin{figure*}[!t]
    \centering
\includegraphics[width=0.85\textwidth]{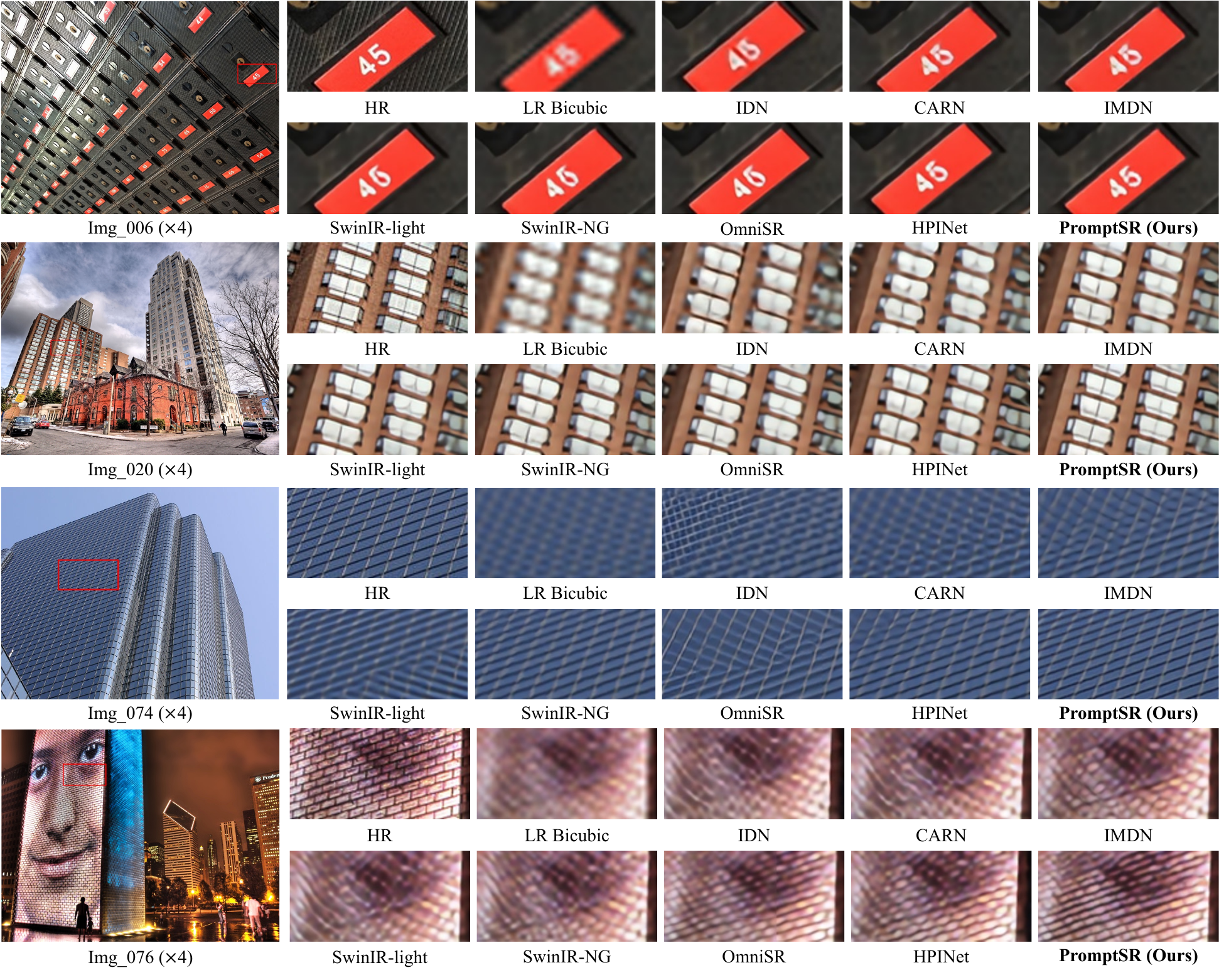}
    \caption{Visual comparison of state-of-the-art lightweight image SR models on Urban100 ($\times$4).}
    % \vspace{-0.1in}
    \label{fig:results} 
\end{figure*}

\begin{figure*}[!t]
    \centering
\includegraphics[width=0.85\textwidth]{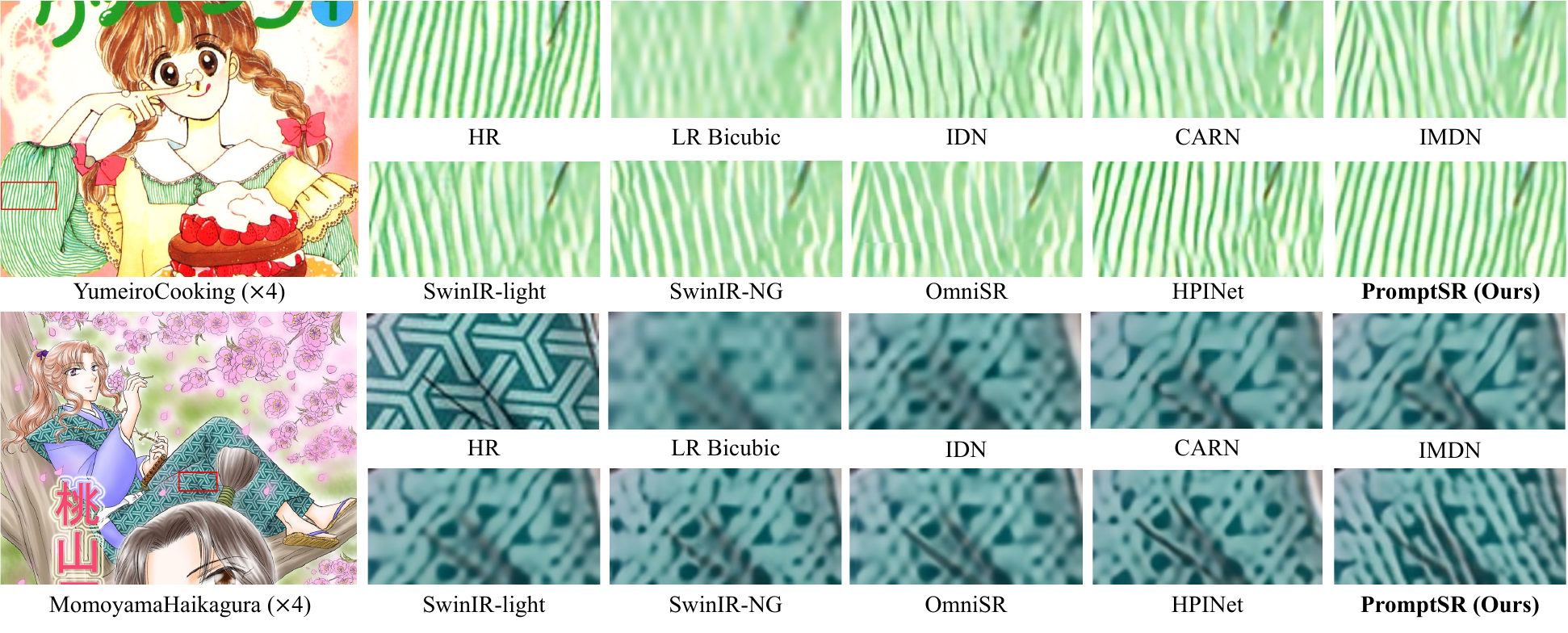}
    \caption{Visual comparison of state-of-the-art lightweight image SR models on Manga109 ($\times$4).}
    \label{fig:results2} 
    \vspace{-0.15in}
\end{figure*}

\section{Experiments}
\subsection{Experimental Settings}
\textit{\textbf{Datasets and Evaluation Metrics.}} We train our model only on DIV2K~\cite{DIV2K} datasets, with 800 images for training, 100 images for testing, and the last 100 images for validation. Following existing lightweight models' comparison~\cite{sun2022shufflemixer, omni_sr, HPI}, the LR images are generated using a bicubic downsampling method. We evaluate the performance of our method on five publicly benchmark test datasets, including Set5~\cite{Set5}, Set14~\cite{Set14}, BSD100~\cite{BSD100}, Urban100~\cite{U100}, and Manga109~\cite{manga109}. 
For quantitative metrics, following~\cite{liang2021swinir, omni_sr, HPI}, we use peak signal-to-noise ratio (PSNR) and structural similarity index (SSIM) on the Y channel of the YCbCr color space to measure the quality of the results.

\textit{\textbf{Implementation Details.}} 
We set the number of the RG as $K=4$, with each RG containing $N=3$ CPBs in the baseline PromptSR. For the enhanced model, PromptSR*, we use $N=4$ CPBs. The downscale ratio of downscaled images $\bm{X}_d$ for getting Anchors $\bm{A}$ is set to $d=8$, and updating weight for APs is set to $\alpha=0.01$. For each CPB, the channel dimension, attention head, and MLP expansion ratio are set as 48, 4, and 1, respectively. The window size in \new{window-based self-attention} (WSA) is set to 16, and the sub-category size in \new{category-based self-attention} (CSA) is set to 128, respectively. Our model is implemented with the PyTorch library with 4 NVIDIA GeForce RTX 3090 GPUs with ADAM optimizer with $\beta_{1}=0.9$, $\beta_{2} = 0.999$, $\epsilon=10^{-8}$. 
We train our model in a batch size 48, using images randomly cropped to $64 \times 64$ pixels. All cropped training images are augmented by random flipping and rotation.
Following previous settings~\cite{liang2021swinir, HAT}, we train the $\times2$ model from scratch for 500K iterations and fine-tune the $\times3$ and $\times4$ models for 250K iterations from the well-trained $\times2$ model. During $\times2$ training, the learning rate is initialized at 5$\times$10$^{-4}$ and reduced by half at iterations [250K,400K,450K,475K,490K]. During $\times3$ and $\times4$ training, the learning rate is initialized at 2$\times$10$^{-4}$ and reduced by half at iterations [150K,200K,225K,240K].
% Following the previous work of \cite{HPI,liang2021swinir}, only the $L_1$ loss is used for training our model.

\subsection{Comparisons with State-of-the-Art Methods}
We compare our proposed method with state-of-the-art lightweight SR models on upscaling factors of $\times$2, $\times$3, and $\times$4, including classical CNN-based models: IDN~\cite{hui2018fast}, CARN~\cite{Ahn_2018_carn}, IMDN~\cite{Hui_2019_imdn}, LAPAR-A~\cite{Li_2020_lapar},  LatticeNet~\cite{Luo_2020_latticenet}, SMFANet$+$~\cite{SMFANet} and latest Transformer/Mamba-based models: SwinIR-light~\cite{liang2021swinir}, ShuffleMixer~\cite{sun2022shufflemixer}, ELAN~\cite{zhang2022efficient}, SwinIR-NG~\cite{Choi_2022_swinirng},  SPIN~\cite{STI}, SRFormer-light~\cite{srformer}, OminiSR~\cite{omni_sr}, HPINet~\cite{HPI}, MambaIR~\cite{mambair}, and ATD-light~\cite{ATD1}.

% SRCNN~\cite{dong2015image}, VDSR~\cite{kim2016accurate}, DRRN~\cite{tai2017image}, A2F~\cite{wang2020lightweight}, MAFFSRN~\cite{muqeet2020multi}, PAN~\cite{zhao2020efficient}, IDN~\cite{hui2018fast}, IMDN~\cite{hui2019lightweight}, LAPAR~\cite{li2020lapar}, RFDN~\cite{liu2020residual}, CARN~\cite{ahn2018fast}, LatticeNet~\cite{luo2020latticenet}, and A-CubeNet~\cite{hang2020attention}, and transformer-based method: SwinIR-light~\cite{liang2021swinir}and ELAN~\cite{zhang2022efficient}.

\begin{figure*}[t]
    \centering
\includegraphics[width=0.9\textwidth]{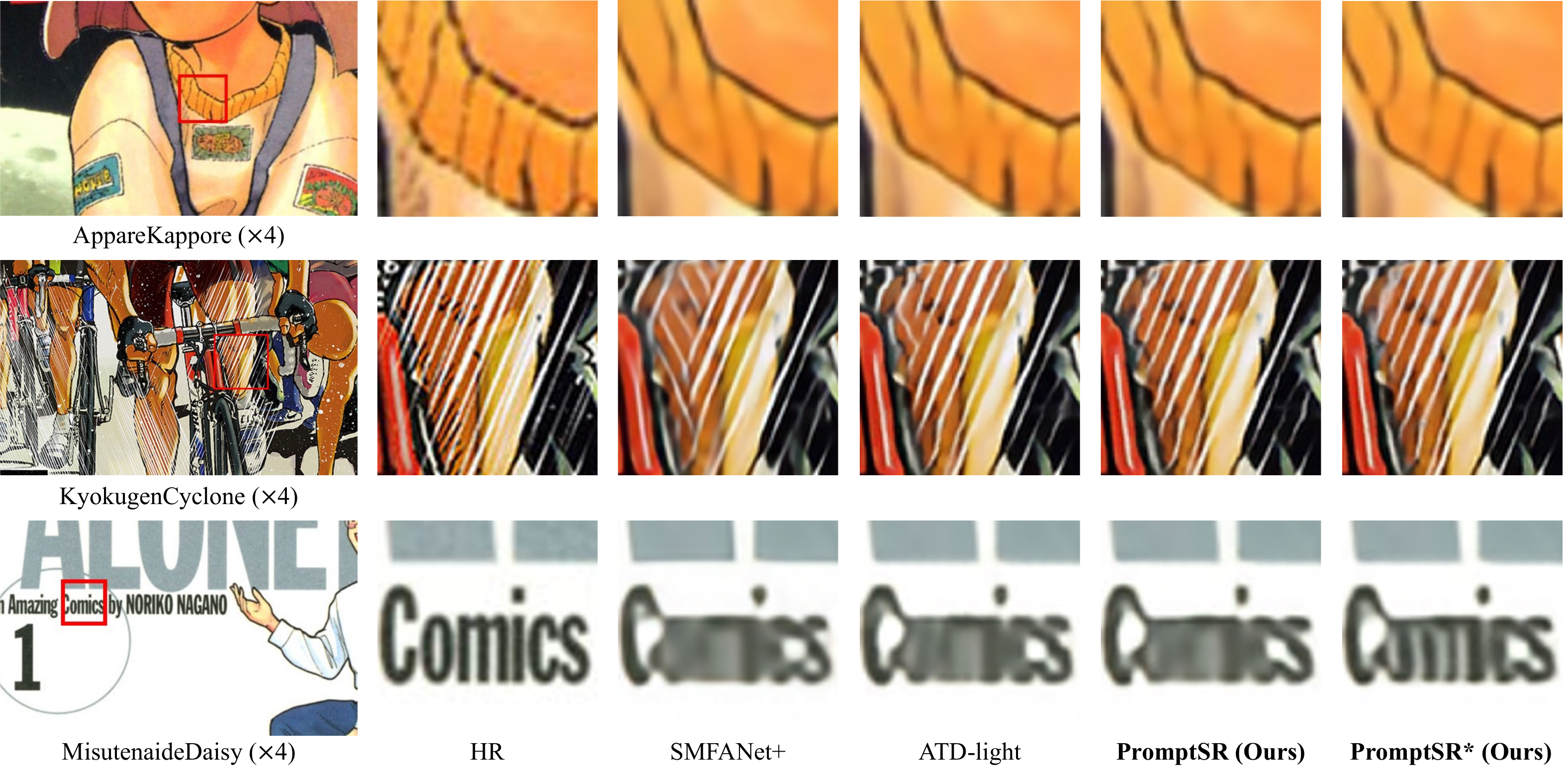}
\caption{Visual comparison of recent state-of-the-art lightweight image SR model on Manga109 ($\times$4).}
    \label{fig:results_comp} 
    \vspace{-0.2in}
\end{figure*}

\textit{\textbf{Quantitative Results.}} The quantitative results for $\times$2, $\times$3, and $\times$4 image \new{SRs} are represented in Table~\ref{tab:results1}. We can observe that our proposed model can consistently outperform existing methods on all five benchmark datasets. 
Compared \new{with} the latest lightweight Transformer-based SR methods OminiSR~\cite{omni_sr}, HPINet~\cite{HPI}, \new{and} MambaIR~\cite{mambair}, our proposed PromptSR method achieves better results with fewer parameters. 
Specifically, PromptSR achieves a significant margin of improvement, with 0.37 dB on Urban100 ($\times4$) and 0.48 dB on \new{the} Manga109 ($\times4$) datasets compared \new{with} OminiSR, and 0.31 dB on \new{the} Urban100 ($\times4$) and 0.31 dB on \new{the} Manga109 ($\times4$) compared \new{with} HPINet, while using less than 780 K parameters. 
Furthermore, even when compared \new{with} the latest state-of-the-art method, ATD-light~\cite{ATD1}, PromptSR achieves better performance in most cases.
Additionally, the enhanced model, PromptSR* significantly outperforms both ATD-light and the baseline model PromptSR, with a small increase in parameters. 
All these improvements \new{demonstrate} the effectiveness of the proposed method.

% Compared with the latest lightweight Transformer-based SR methods OminiSR~\cite{omni_sr} and HPINet~\cite{HPI}, our proposed HCSA method can achieve better results with less parameter size. Specifically, the HCSA gains a large margin of 0.38dB on $\times4$ Urban100 and 0.48dB on the $\times4$ Manga109 than OminiSR, and 0.32dB on $\times4$ Urban100 and 0.31dB on$ \times4$ Manga109 than HPINet.

\begin{figure*}[t]
    \centering
\includegraphics[width=0.9\textwidth]{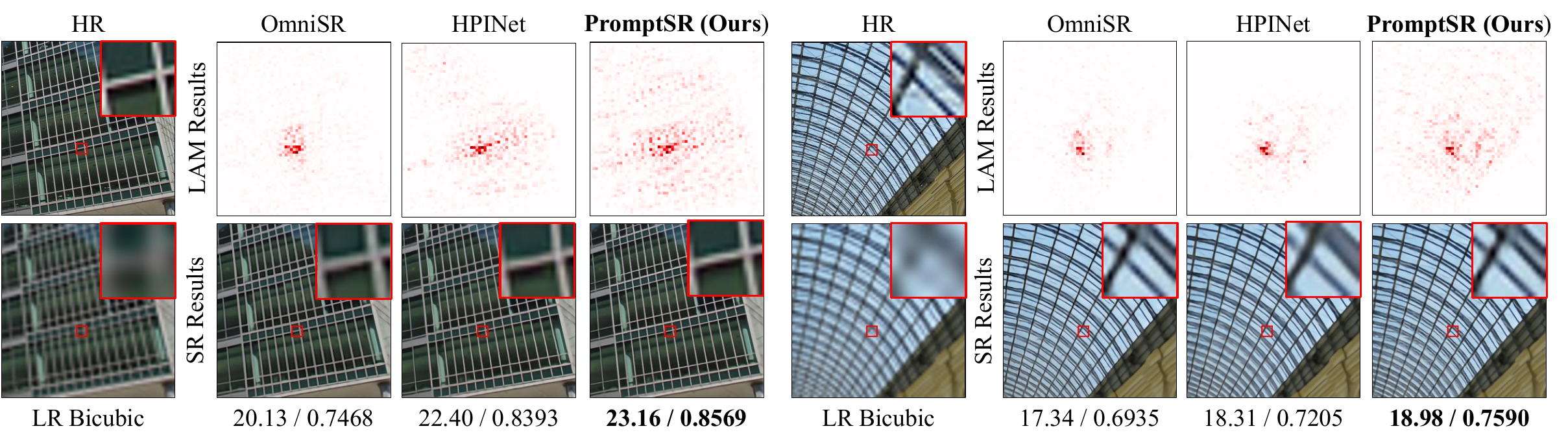}
    \caption{Visual comparison of LAM and SR results on Urban100 ($\times$4) between OmniSR, HPINet, and our PromptSR. }
      \vspace{-0.1in}
    \label{fig:LAM_results_u} 
\end{figure*}

\begin{figure*}[!t]
    \centering
\includegraphics[width=0.9\textwidth]{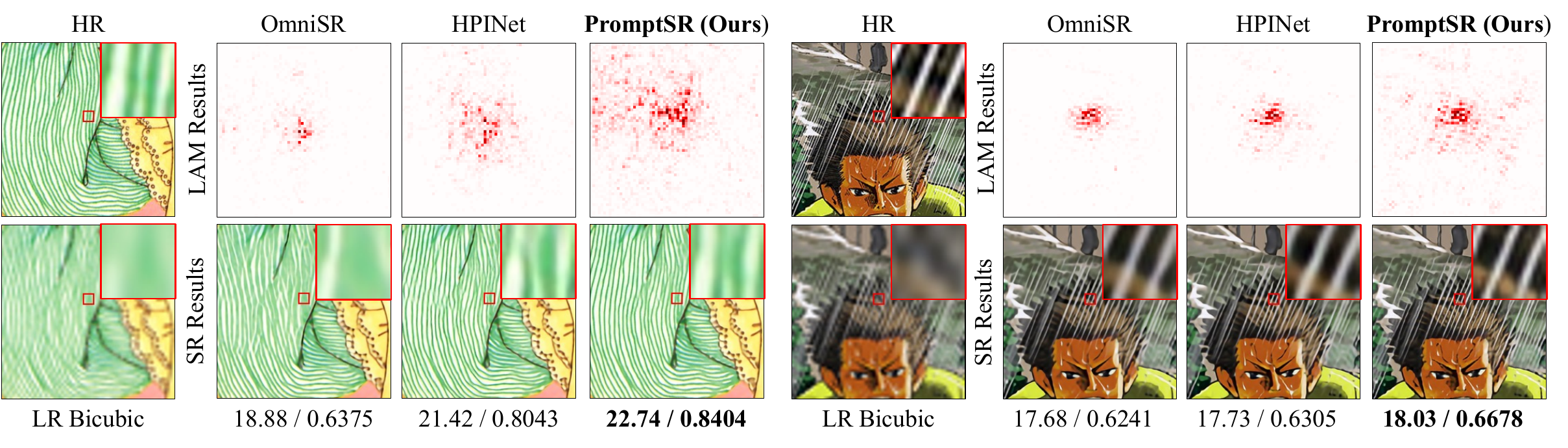}
    % \vspace{-0.2in}
    \caption{Visual comparison of LAM and SR results on Manga109 ($\times$4) between OmniSR, HPINet, and our PromptSR.}
          \vspace{-0.1in}
    \label{fig:LAM_results_m} 
\end{figure*}

\textit{\textbf{Qualitative Results.}} The qualitative visual comparisons on \new{the} Urban100 ($\times$4) and Manga109 ($\times$4) datasets are given in Fig.~\ref{fig:results} and Fig.~\ref{fig:results2}, respectively. We can observe that most of the compared methods \new{cannot effectively} recover accurate textures and suffer from severe blurring artifacts in some challenging cases such as ''image\_076'' from the Urban100 dataset. Even the latest lightweight Transformer-based SR methods OminiSR~\cite{omni_sr} and HPINet~\cite{HPI} may generate inconsistent structural details \new{because of} the limited receptive field. 
In contrast, PromptSR effectively captures global priors through the global anchor prompting layer (GAPL) to restore consistent and sharp textures while alleviating blurring artifacts through the coarse-to-fine local prompting layers (LPLs) to refine details.
% For example, in ''image\_006'', the most compared methods recover a blurry result of the number '45', which may result in it being falsely regarded as '46', while our PromptSR can recover the correct number details. 
% For instance in the Urban100 dataset, in ''image\_006,'' most of the compared methods recover a blurry result of the number "45", which could be mistakenly interpreted as "46". In contrast, our PromptSR method accurately recovers the detailed features of the correct number.
As shown in Fig.~\ref{fig:results} of the Urban100 dataset, on "img\_006", most of the compared methods produce blurry results for the number "45" which could be mistakenly interpreted as "46". In contrast, our PromptSR method can accurately recover the details of the number, preventing this misinterpretation.
Additionally, on "img\_020", "img\_074" and "img\_076", our PromptSR can produce more accurate and sharp lines, enabling the generation of consistent structural details, such as mesh patterns. 
\new{Moreover, most of the} other compared methods fail to recover these details and may produce undesired artifacts.
Similar behaviors are also observed in the Manga109 dataset as shown in Fig.~\ref{fig:results2}. 
For example, on "YumeiroCooking", our PromptSR can recover the most accurate and clear striped texture details \new{compared with the} other methods.
However, there are also some failure cases where all models struggle to generate precise texture details, such as the pattern shape of the cloth in “MomoyamaHaikagura”. 
Nevertheless, PromptSR still manages to recover more consistent details than \new{the} other methods \new{do}.
These qualitative results highlight the efficacy of our PromptSR, which benefits from our proposed prompting strategies to enable an enlarged global receptive field.

\begin{figure}[t]
    \centering
\includegraphics[width=0.47\textwidth]{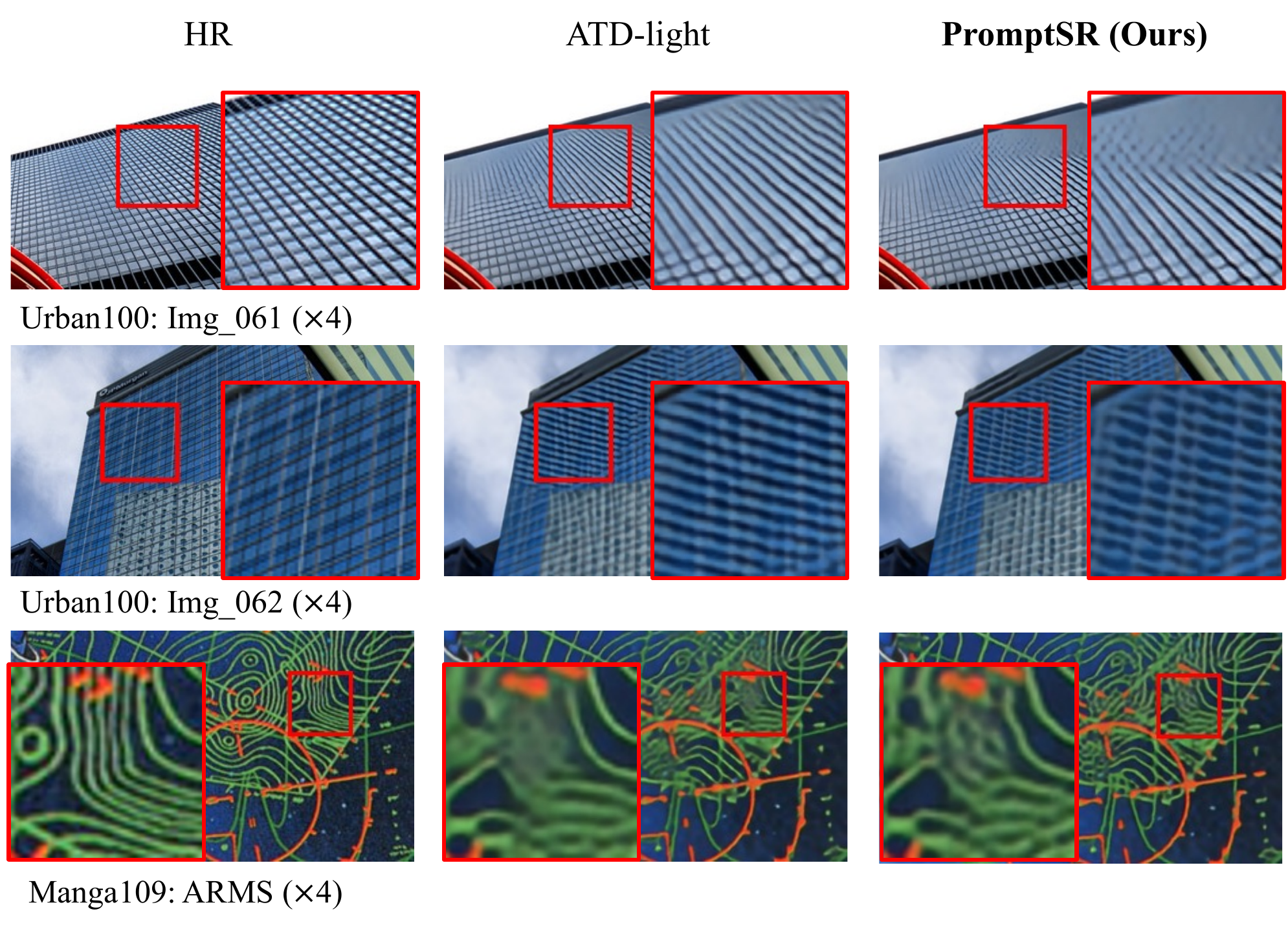}
\caption{Failure cases on Urban100 ($\times$4) and Manga109 ($\times$4).}
    \label{fig:results_fail} 
    \vspace{-0.2in}
\end{figure}

\textit{\textbf{Qualitative Comparison with More Recent Approaches.}} We further provide qualitative comparisons with more recent approaches, such as SMFANet$+$~\cite{SMFANet} and ATD-light~\cite{ATD1}, on \new{several} more challenging scenarios. The visual comparisons, including clothing textures (e.g., collars), raindrop details, and English text, are presented in Figure \ref{fig:results_comp}. As illustrated, \new{although} all methods face difficulties in accurately recovering the contents, our approach exhibits better fidelity in preserving intricate edge details, such as lines and curves. This highlights the robustness of our proposed GAPL, which leverages the inherent cross-scale similarity property observed in natural images, and the effectiveness of our prompting strategies in handling these challenging scenarios. Furthermore, our enhanced model, PromptSR*, achieves even better reconstruction of fine details, delivering the best performance among the compared methods.

\begin{table}[b]
    \centering
    \caption{Theoretical time and space complexity comparison between vanilla self-attention (SA), window-based self-attention (WSA), and our proposed Global Anchor Prompting Layer (GAPL), where $H, W$ and $C$ represent the input features' size and channel and $H_d$ and $W_d$ represents window size/the downscaled size of ours.}
     \setlength{\tabcolsep}{1mm}
    \begin{tabular}{lcc}
        \toprule
        \textbf{Module} & \textbf{Time Complexity} & \textbf{Space Complexity}  \\ 
        \hline
        SA~\cite{dosovitskiy2020image} & $\mathcal{O}(HWC^2 + H^2W^2C)$ & $\mathcal{O}(HWC + H^2W^2)$  \\ 
        
        WSA~\cite{liang2021swinir} & $\mathcal{O}(HWC^2 + HWH_dW_dC)$ & $\mathcal{O}(HWC + HWH_dW_d)$   \\
            \rowcolor{Gray}
        \textbf{GAPL (Ours)} & $\mathcal{O}(HWC^2 + HWH_dW_dC)$ & $\mathcal{O}(HWC+HWH_dW_d)$\\
    \bottomrule
    \end{tabular}
    \label{tab:complex}
\end{table}
 
\begin{table*}[]
    \centering
\caption{Parameters and computational costs comparison of existing advanced large SR models and state-of-the-art lightweight SR models on Urban100 ($\times$4) and on Manga109 ($\times$4). The best results are highlighted in \textbf{bold}.}
\setlength{\tabcolsep}{3mm}
\resizebox{0.77\textwidth}{!}{
\begin{tabular}{l|c|c|c|cc|cc}
\bottomrule
  \multirow{2}{*}{\textbf{Method}} &   \multirow{2}{*}{\textbf{Scale}} &   \multirow{2}{*}{\textbf{\#Param.}} &    \multirow{2}{*}{\textbf{\#Multi-Adds.}} & \multicolumn{2}{c}{\textbf{Urban100}} & \multicolumn{2}{|c}{\textbf{Manga109}} \\ \cline{5-8}
 & & & & PSNR & SSIM & PSNR & SSIM  \\ 
\hline
RCAN~\cite{zhang2018image} & $\times$4 & 15.6M & 918G & 26.82 & 0.8087  & 31.22 & 0.917\\ 
SAN~\cite{dai2019second} & $\times$4 & 15.9M & 937G & 26.79 & 0.8068 & 31.18 & 0.9169 \\
NLSN~\cite{NLSN} & $\times$4 & 44.2M & 2956G & 26.96 & 0.8109 & 31.27 & 0.9184\\  
SwinIR~\cite{liang2021swinir} & $\times$4 & 11.9M & 584G & \textbf{27.07} & \textbf{0.8164} & \textbf{31.67} & \textbf{0.9226} \\  
\hline

SwinIR-light~\cite{liang2021swinir} & $\times$4 & 930K & 49.6G & 26.47 & 0.7980 & 30.92 & 0.9151 \\

SRFormer-light~\cite{srformer} & $\times$4 & 873K & 62.8G & 26.67 & 0.8032 & 31.17 & 0.9165\\

OmniSR~\cite{omni_sr} & $\times$4 & 792K & 45.3G  & 26.65 & 0.8018 & 31.02 & 0.9151\\

HPINet~\cite{HPI} & $\times$4 & 895K & 81.0G  & 26.71 & 0.8043 & 31.19 & 0.9162\\
    \rowcolor{Gray}
\textbf{PromptSR (Ours)} &$\times$4& 779K &  53.5G & \textbf{27.02} & \textbf{0.8116} & \textbf{31.50} & \textbf{0.9198}\\

\toprule
\end{tabular}}
\vspace{-0.15in}
\label{tab:complex1}
\end{table*}

\textit{\textbf{LAM Results.}} We further utilize \new{the} LAM~\cite{gu2021interpreting} as a tool to visualize the most contributing pixels regarding the SR results of the selected regions. The visual comparisons and the SR results on Urban100 ($\times$4) and on Manga109 ($\times$4) are presented in Fig.~\ref{fig:LAM_results_u} and Fig.~\ref{fig:LAM_results_m}, respectively. As we can see, our PromptSR \new{method} activates more pixels than \new{the} state-of-the-art methods OmniSR~\cite{omni_sr} and HPINet~\cite{HPI}, demonstrating its effectiveness in capturing long-range interactions through the proposed prompting strategies.
Consequently, PromptSR achieves more accurate textures and higher PSNR and SSIM \new{values than} these state-of-the-art methods \new{do}.
% As a result, our PromptSR achieves more accurate textures and higher PSNR and SSIM than these methods.

\textit{\textbf{Failure Case Analysis.}} 
We further \new{present} some failure cases in  Fig.~\ref{fig:results_fail} and provide a detailed analysis.
Despite the overall effectiveness of the proposed PromptSR, there are scenarios where the model struggles to deliver satisfactory results and performs worse than the state-of-the-art model, ATD-light~\cite{ATD1}. The experimental results indicate that image SR becomes particularly challenging for all methods in cases with highly detailed and complex textures. While our approach excels at recovering missing details, it can occasionally produce larger artifacts in these intricate regions, as the proposed coarse-to-fine prompting module tends to aggressively model and reconstruct the missing patterns, e.g., the finely woven network \new{presented} in the “ARMS” presented.

\subsection{Model Complexity Analysis}
To demonstrate the effectiveness and efficiency of the proposed GAPL and PromptSR, we first \new{present} a theoretical analysis of GAPL, and then evaluate our PromptSR with state-of-the-art SR methods on the Urban100~\cite{U100} dataset. 

\textit{\textbf{Theoretical Analysis.}}
Tabel~\ref{tab:complex} provides a detailed comparison of the time and space complexity upper bound between vanilla self-attention (SA)~\cite{dosovitskiy2020image}, window-based self-attention (WSA)~\cite{liang2021swinir}, and our proposed global anchor prompting layer (GAPL). 
We assume \new{that} the window size in \new{the} WSA and the size of our downscaled feature $X_d$ are equal, represented as $H_d, W_d$. 
Since they are significantly smaller than the size of the original input features, i.e., $H_d \ll H$ and $W_d \ll W$, the complexity of our model is comparable to that of WSA \new{but} significantly lower than that of vanilla SA. 
This comparison reveals that our GAPL approach can maintain a lightweight structure similar to WSA while achieving a larger global receptive field, demonstrating the effectiveness of our PromptSR.

% The average running time is evaluated on the SOTS-indoor dataset, FLOPs are obtained based on 256$\times$256 resolution image patch, and each result is acquired by repeating the experiment five times to guarantee a fair comparison. From Figure~\ref{fig:complexity}, we can easily conclude that our MITNet obtains the best performance with the fewest number of parameters, the fewest FLOPs, and the second-fast inference time. Thus, our MITNet achieves an excellent model complexity versus performance trade-off. FSDGN~\cite{fsdgn} utilizes dual-domain information while having a noticeable performance gap with ours, which shows it is necessary for adequate feature interaction and redundant feature removal.

% \begin{figure}[t]
%     \centering
%     \includegraphics[width=0.45\textwidth]{fig/tsne.pdf}
%     \caption{The t-SNE visualization of Anchor Prompts in 4 Residual Groups. Images are randomly selected in 10 categories from ImageNet~\cite{krizhevsky2012imagenet} with 5000 samples.}
%     \label{fig:tsne}
% \end{figure}

\textit{\textbf{Model Complexity Comparison.}} We compare our PromptSR with state-of-the-art lightweight SR models in terms of parameters (Param.), multiply-add operations (Multi-Adds), PSNR, and SSIM. 
To provide a comprehensive comparison, we also include advanced large models such as \new{the} classical CNN model RCAN~\cite{zhang2018image}, two CNN models equipped with non-local attention, i.e., SAN~\cite{dai2019second} and NLSN~\cite{NLSN}, and the well-known Transformer-based model SwinIR~\cite{liang2021swinir} in Table\ref{tab:complex1}. 
We evaluate all models on the Urban100 ($\times4$) and Manga109 ($\times4$) datasets, with Multi-Add operations counted for a given $1280 \times 720$ HR image, following previous settings~\cite{srformer,HPI}. Our PromptSR achieves the best PSNR and SSIM performance with the fewest parameters, \new{although} the \new{computational} cost is slightly higher than \new{that of} SwinIR-light~\cite{liang2021swinir} and OmniSR~\cite{omni_sr}. 
Compared with existing advanced large models, our PromptSR outperforms RCAN, SAN, and NLSN by a significant margin. Additionally, our PromptSR \new{performs comparably} to SwinIR~\cite{liang2021swinir} (11.9 M parameters) with \new{fewer than} 0.78 M parameters.

% \subsection{Anchor Prompts t-SNE visualization}
% To validate that our introduced Anchor Prompts $\em{P}$ can promotes to preserve the most crucial features and the effectiveness of 

% the discriminator to perform more fine-grained discrimination, 

\begin{figure}[t]
    \centering    \includegraphics[width=0.47\textwidth]{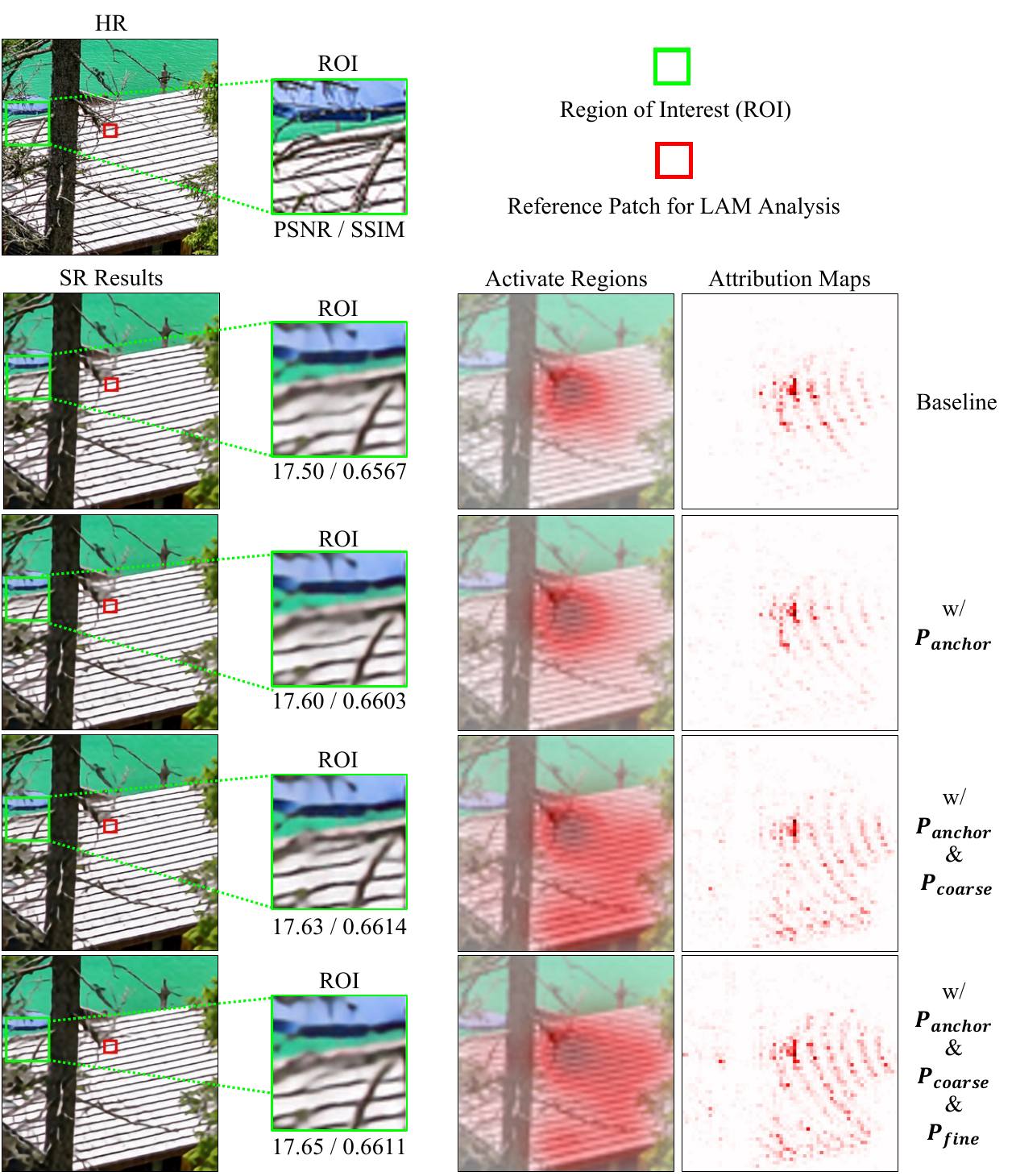}
    \caption{Ablation study on the effects of our proposed three prompting strategies. Visualize the corresponding activated regions through LAM analysis.}
    \label{fig:abla_vs}
        \vspace{-0.1in}
\end{figure}

\subsection{Ablation Studies}
We further conduct ablation studies to evaluate the proposed components in our PromptSR. 
For fair comparisons with the baselines, we implement all experiments on the rescaled PromptSR Model ($\times 4$) with each \new{residual group} (RG) containing $N = 2$ \new{cascade prompting blocks} (CPBs).
We train all \new{the} models for 250k iterations on the DIV2K~\cite{DIV2K} dataset with halved learning rates and evaluate them on \new{the} Urban100~\cite{U100} and Manga109~\cite{manga109} datasets.

\textit{\textbf{Effects of Prompting Layers.}} To evaluate the effectiveness of our proposed three prompting strategies, i.e., anchor prompting, coarse prompting, and fine prompting, we construct four different models and \new{compare} the quantitative results in Table~\ref{tab:abla1}. The first model serves as the baseline and \new{is constructed} by removing the \new{global anchor prompting layer} (GAPL) and \new{local coarse-to-fine prompting layers} (LPLs), thereby disabling all three promptings. For a fair comparison, we replace these layers with window-based self-attention (WSA) layers. Table~\ref{tab:abla1} shows that the baseline model achieves the worst results, demonstrating the limited receptive fields imposed by window partitioning. The second model adopts the proposed GAPL and leverages the constructed anchor prompts $\bm{P}$ to introduce global priors, outperforming the baseline model by \new{approximately} 0.1 dB on our evaluated datasets. In the third model, we adopt both \new{the} GAPL and LPL layers but replace the fine prompts with the coarse prompts by changing the input $\bm{M}_{fine}$ to $\bm{M}_{coarse}$. With single coarse prompts, this model gains a further 0.2 dB improvement on \new{the} Urban100. Finally, the last model, \new{which is} equipped with global anchor prompting and local coarse-to-fine prompting strategies, achieves the best performance. 

\begin{table}[]
    \centering
    \caption{Ablation study on the effects of three promptings on Urban100 ($\times$4) and Manga109 ($\times$4). }
   \begin{tabular}{ccccccc}
    \toprule
\multicolumn{3}{c}{\textbf{Prompting}} & \multicolumn{2}{c}{\textbf{Urban100}} & \multicolumn{2}{c}{\textbf{Manga109}} \\
        $\bm{P}_{anchor}$ & $\bm{P}_{coarse}$ & $\bm{P}_{fine}$ &  PSNR & SSIM & PSNR & SSIM   \\ \midrule
                &         &         &  26.49 & 0.7979 &  30.87 & 0.9136  \\
        \cmark  &         &         &  26.58 & 0.7999  & 30.92 & 0.9140  \\
        % \cmark  & \cmark  &          &  26.60  & 0.7999 & 30.92 & 0.9140 \\
        \cmark  & \cmark  &          &  26.59  & 0.7997 & 30.95 & 0.9143 \\
            \rowcolor{Gray}
        \cmark &  \cmark  & \cmark   &  \textbf{26.63}  & \textbf{0.8006}  & \textbf{30.98} & \textbf{0.9146}  \\
 \bottomrule
        %0.7399
    \end{tabular}
    \label{tab:abla1}
 % \vspace{-0.2in}
\end{table}
\begin{table}[t]
    \centering
      \caption{Ablation study on the effects of the downscale size $d$ on Urban100 ($\times$4) and Manga109 ($\times$4). }
      \setlength{\tabcolsep}{2mm}
    \begin{tabular}{ccccccc}
            \toprule
           \multirow{2}{*}{$\bm{d}$}  &  \multirow{2}{*}{\textbf{\#Params.}}  & \multirow{2}{*}{\textbf{\#Multi-Adds.}} & \multicolumn{2}{c}{\textbf{Urban100}} & \multicolumn{2}{c}{\textbf{Manga109}} \\
           & & &  PSNR & SSIM & PSNR & SSIM   \\    \midrule
                2  & 534K & 81.1G & 26.34 & 0.7920 & 30.59 & 0.9100  \\
                4   & 540K & 46.8G & 26.61 & 0.8000 & 30.89 & 0.9140  \\
                \rowcolor{Gray}
                8   & 562K & 38.3G & \textbf{26.63} & \textbf{0.8006} & \textbf{30.98} & \textbf{0.9146} \\
        \bottomrule
        \end{tabular}
        \vspace{-0.15in}
    \label{tab:abla2}
\end{table}

\begin{table}[]
    \centering
      \caption{Ablation study on the effects of the weighting factor $\alpha$ on Manga109 ($\times$4). }
      \setlength{\tabcolsep}{2mm}
    \begin{tabular}{>{\centering\arraybackslash}p{1.5cm}cccc}
            \toprule
          \multirow{2}{*}{$\bm{\alpha}$}   & \multicolumn{2}{c}{\textbf{Urban100}} & \multicolumn{2}{c}{\textbf{Manga109}} \\
           &   PSNR & SSIM & PSNR & SSIM   \\
             \midrule
                0  &  26.62 & 0.8001 & 30.86  & 0.9135 \\
                0.1   & \textbf{26.64} & \textbf{0.8009} & 30.94 & 0.9142   \\
                \rowcolor{Gray}
                0.01   & 26.63 & 0.8006 & \textbf{30.98} & \textbf{0.9146}  \\
        \bottomrule
        \end{tabular}
    \label{tab:abla3}
\end{table}
\begin{table}[]
    \centering
      \caption{Ablation study on the effects of the window size $ws$ on Urban100 ($\times$4).}
      \setlength{\tabcolsep}{2mm}
    \begin{tabular}{>{\centering\arraybackslash}p{1.5cm}cccc}
            \toprule
          $\bm{ws}$  &  \textbf{\#Params.}  & \textbf{\#Multi-Adds.} & \textbf{PSNR} & \textbf{SSIM} \\
             \midrule
                (8, 8)  & 562K & 34.4G & 26.32 & 0.7936  \\
                \rowcolor{Gray}
                (16, 16) & 562K & 38.3G & \textbf{26.63} &\textbf{0.8006}   \\
        \bottomrule
        \end{tabular}
        \vspace{-0.15in}
    \label{tab:abla4}
\end{table} 

We also provide qualitative visual comparisons of \new{the} LAM results, including both \new{the} activated regions and attribution maps on Urban100 ($\times$4) in Fig.~\ref{fig:abla_vs}. With our proposed three prompting strategies, we observe that more pixels are gradually activated for the same region in SR, resulting in enlarged receptive fields.
As a result, the corresponding SR results achieve higher PSNR and SSIM \new{values} and have better visual quality \new{than} the baseline methods \new{do}.
Both \new{the} quantitative and qualitative comparisons demonstrate the effectiveness and superiority of our proposed three prompting strategies.

% a wider range of receptive fields for each prompt,

\textit{\textbf{Effects of Downscale Ratio $\bm{d}$.}} In our proposed GAPL, the downscale ratio $\bm{d}$ determines the number ($HW/d^2$) of \new{the} constructed anchor prompts (APs). 
A larger downscale ratio $d$ results in fewer APs, potentially facilitating each single AP \new{to focus better} on the most important global priors and reducing attention to capture details and redundant information for each prompt. 
Consequently, our PromptSR may benefit from a larger downscale ratio $d$.
To evaluate the influence of varying downscale ratios, we conducted experiments ranging downscale ratios $d$ from 2 to 8. 
The experimental results are shown in Table~\ref{tab:abla2}. 
We find that the model achieves the best results at $d = 8$, demonstrating the effectiveness of our proposed APs.
Moreover, we observe that a larger downscale ratio $d$ also reduces computational overhead \new{because} fewer APs \new{participate} in the calculation.
Hence, we set the downscale ratio $d = 8$ for our final model.

\textit{\textbf{Effects of the Weighting Factor $\bm{\alpha}$.}} In our proposed GAPL, the weighting factor $\alpha$ plays a critical role in balancing the proportion of retaining the APs from the previous stage and generating a new one for the current stage. To evaluate its impact, we conducted a series of experiments by varying the value of $\alpha$ from 0 to 0.1. The results, shown in Table~\ref{tab:abla4}, demonstrate that the model without retaining the APs from the previous stage (i.e., $\alpha = 0$) resulted in a significant drop in performance, highlighting the importance of leveraging previous prompts to inform the current stage. As the value of $\alpha$ increased from 0.01 to 0.1, the results fluctuated: the model performed better \new{when} the model \new{behaved} better on Urban100 ($\times 4$) with $\alpha = 0.1$ but worse on Manga109 ($\times 4$). For our final model, we set $\alpha = 0.01$ to achieve a balanced performance across datasets.

\textit{\textbf{Effects of the Window Size $\bm{ws}$.}} In our proposed LPLs, the window size plays a crucial role in the employed WSA module. 
Larger window sizes typically enable the self-attention mechanism to capture a broader range of dependencies. To evaluate its effectiveness, we conduct experiments varying the window size. As shown in Table~\ref{tab:abla4}, the model with a larger window size of 16$\times$16 achieves better performance, especially \new{for} Urban100 ($\times$4), \new{although} it comes at the cost of increased computational requirements. These results demonstrate that enlarging the window size can effectively enhance our model performance. To balance computational efficiency and performance, we set $ws=16$ for our final model.

\subsection{T-SNE visualization of Anchor Prompts}
\begin{figure*}[t]
    \centering
\includegraphics[width=0.97\textwidth]{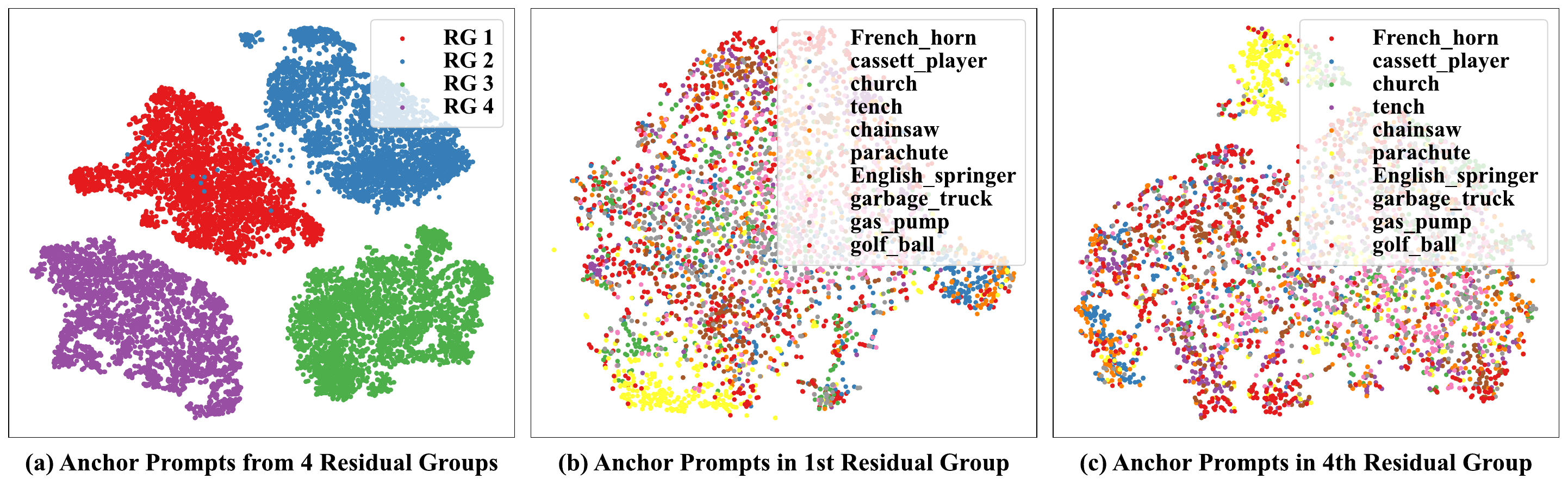}
    \caption{The t-SNE visualization of Anchor Prompts: images are randomly selected in 10 categories from ImageNet~\cite{krizhevsky2012imagenet} with a total of 5000 image samples. (a) Collect Anchor Prompts from 4 Residual Groups (RGs)  $\{\bm{P}^1,\bm{P}^2,\bm{P}^3,\bm{P}^4\}$ together and visualize them. (b) Visualize the Anchor Prompts $\bm{P}^1$ individually to explore the semantics. (c) Visualize the Anchor Prompts $\bm{P}^4$ individually to explore the semantics.}
    \label{fig:tsne}
    \vspace{-0.15in}
\end{figure*}

\textit{\textbf{Analytics of Anchor Prompts.}} To validate the effectiveness of our introduced APs and understand the rationale behind our proposed \textit{\textbf{Anchor Prompt Update}} within each \new{residual group} (RG), 
we visualize the intermediate $\bm{P}$ of APs from 4 RGs with t-SNE in our PromptSR. We randomly select \new{approximately} 5000 image samples in 10 categories from ImageNet~\cite{krizhevsky2012imagenet}. 
The LR images are first generated using a bicubic downsampling method and then sent to our models to obtain $\bm{P}^1,\bm{P}^2,\bm{P}^3$, and $\bm{P}^4$.
Each $\bm{P}^i$ is extracted from the last \new{cascade prompting block} (CPB) within each RG for a fair comparison. We then collect $\{\bm{P}^1,\bm{P}^2,\bm{P}^3,\bm{P}^4\}$ together and visualize them in 2D dimensions by using t-SNE.
As shown in Fig.~\ref{fig:tsne}(a), $\bm{P}^i$ from 4 RGs are well-clustered into 4 categories, indicating that $\bm{P}^i$ from different RGs exhibit distinct characteristics.
These results highlight that although $\bm{P}$ \new{is} designed to capture consistent global priors, PromptSR enables $\bm{P}^i$ at different depths (RGs) to capture various global features, thereby facilitating $\bm{P}^i$ with enhanced comprehensive representations.
Furthermore, these results also demonstrate the effectiveness of our introduced \textit{\textbf{Anchor Prompt Update}}, which enables the update within each RG while avoiding cross-RG updates, thereby preserving the diversity of $\bm{P}^i$ across different RGs.

% \begin{figure}[t]
%     \centering
% \includegraphics[width=0.42\textwidth]{fig/tsne.pdf}
%     \caption{The t-SNE visualization of Anchor Prompts from 4 Residual Groups. Images are randomly selected in 10 categories from ImageNet~\cite{krizhevsky2012imagenet} with a total of 5000 image samples.}
%     \label{fig:tsne}
% \end{figure}

% \begin{figure}[t]
%     \centering
% \includegraphics[width=0.41\textwidth]{fig/tsne_compare1.pdf}
%     \caption{The t-SNE visualization of Anchor Prompts of RG1 and RG4, respectively. Images are randomly selected in 10 categories from ImageNet~\cite{krizhevsky2012imagenet} with a total of 5000 image samples.}
%     \label{fig:tsne1}
% \end{figure}

\textit{\textbf{Limitations of Anchor Prompts.}} 
Recently, several studies~\cite{aakerberg2022semantic, wu2024seesr} have considered semantic information as crucial priors for achieving image SR, demonstrating good performance.
This motivates us to explore whether our constructed APs preserve important semantic information so that samples from different image categories can be distinguished through APs. 
To investigate this, we use t-SNE to visualize $\bm{P}^i$ within a single RG, as shown in Fig.~\ref{fig:tsne}.
Specifically, Fig.~\ref{fig:tsne}(b) and Fig.~\ref{fig:tsne}(c) provide the visualizations of $\bm{P}^1$ and $\bm{P}^4$, respectively. 
The results indicate that neither of them is well-clustered with respect to the image categories. This observation suggests that our constructed APs, \new{which are} designed to filter out less important local details, may also filter out some important semantic information, which could be detrimental to performance. 
In the future, we plan to explore other techniques to preserve this semantic information to further improve the performance of image SR.

\section{Conclusion}

%In this work, we propose PromptSR, a prompting-empowered lightweight and high-performance image SR method, where the key insight lies in leveraging low-dimensional features as anchors to provide prompt for SR. 
In this paper, we introduce PromptSR, a lightweight and high-performance image SR method that harnesses the power of prompting. 
The key insight of PromptSR is the construction of \new{anchor prompts} (APs) with enhanced global perception to enlarge the receptive field.
PromptSR consists of a series of \new{cascade prompting blocks} (CPB) cascaded by a \new{global anchor prompting layer} (GAPL) and two \new{local prompting layers} (LPLs), designed to promote global information access and local information refinement
The GAPL capitalizes on the cross-scale similarity property to construct low-dimensional APs that can freely aggregate global information from the original image space and then utilizes these APs to provide global prompts.
% These APs are then utilized to provide global prompts, facilitating long-range token connections.
Experiments demonstrated the effectiveness of our constructed APs, which can capture consistent global priors from different inputs. The two LPLs \new{subsequently} combine category-based and window-based self-attention mechanisms to refine the representation in a coarse-to-fine manner, which leverages attention maps from the GAPL as additional global prompts to guide the refining process.
Extensive experiments on five public benchmarks validate the superiority of our PromptSR, \new{which outperforms} state-of-the-art lightweight SR methods in quantitative, qualitative, and complexity evaluations.

\bibliographystyle{ieeetr}

\bibliography{ref}

\begin{IEEEbiography}[{\includegraphics[width=1in,height=1.25in,clip,keepaspectratio]{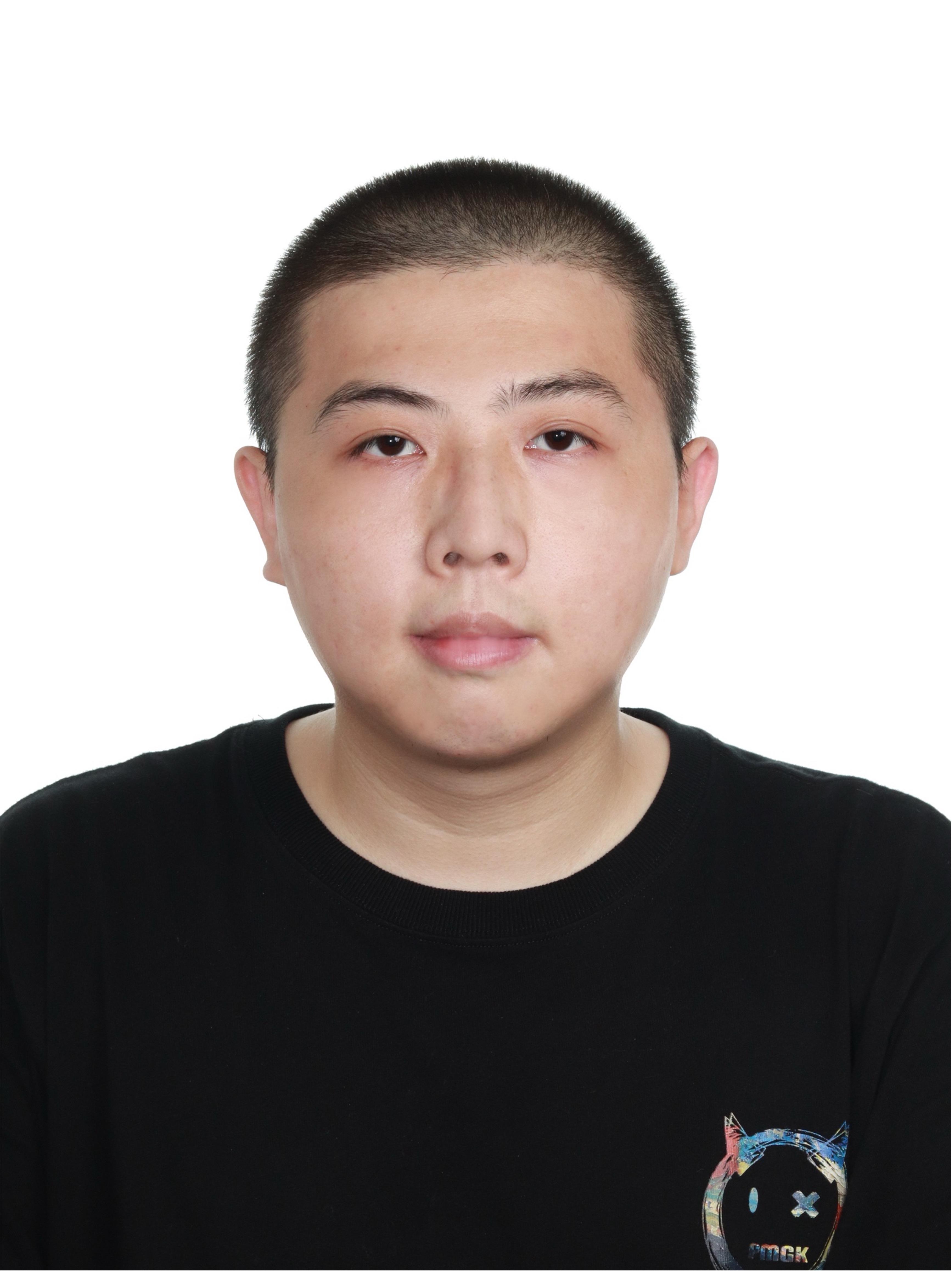}}]{Wenyang Liu} (Student Member, IEEE) received BEng and MEng degrees in the College of Computer Science from Chongqing University, Chongqing, China, in 2017 and 2020, respectively. He is currently pursuing the Ph.D. degree in the School of Electrical and Electronic Engineering from Nanyang Technological University, Singapore. His research interests include computer vision and digital forensics, with a focus on image restoration, super-resolution, and generative models.
\end{IEEEbiography}

\begin{IEEEbiography}[{\includegraphics[width=1in,height=1.25in,clip,keepaspectratio]{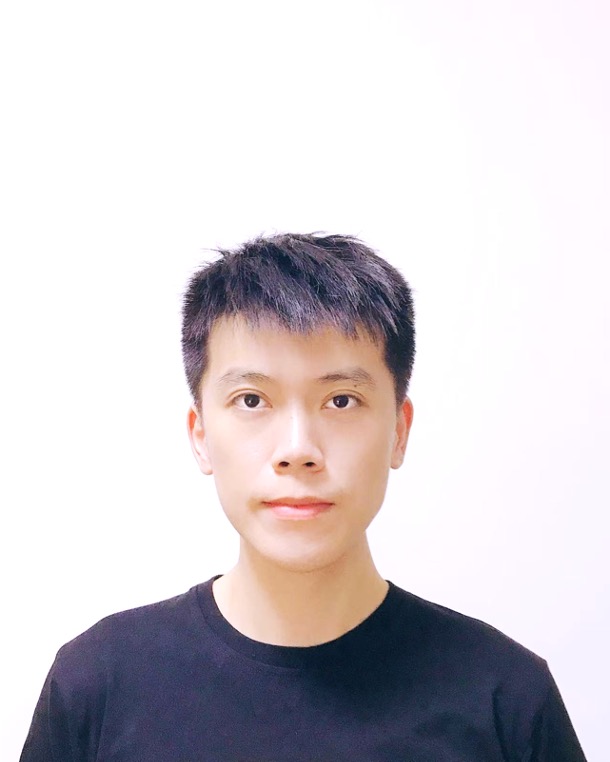}}]{Chen Cai}received his Ph.D. degree from the School of Electrical and Electronic Engineering, Nanyang Technological University, Singapore, in 2024, and the B.S. degree in Electrical and Computer Engineering from the National University of Singapore. He is currently a Research Scientist at the Centre for Remote Imaging, Sensing and Processing (CRISP), National University of Singapore. His research interests include computer vision, NLP, and multimodal learning.
\end{IEEEbiography}

\begin{IEEEbiography}[{\includegraphics[width=1in,height=1.25in,clip,keepaspectratio]{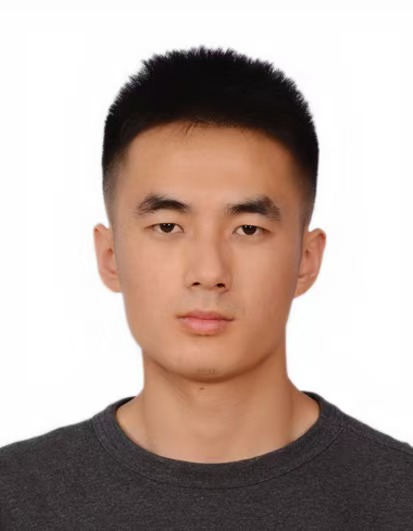}}]{Jianjun Gao} received the BEng degree in Electronic Information Science and Technology from Sichuan University, Chengdu, China, in 2020 and the M.Sc. degree in Communication Engineering from Nanyang Technological University, Singapore, in 2021. He is currently working toward the Ph.D. degree at the School of Electrical and Electronic Engineering, Nanyang Technological University, Singapore. His research interests include computer vision, artificial intelligence, and image/video processing.
\end{IEEEbiography}

\begin{IEEEbiography}[{\includegraphics[width=1in,height=1.25in,clip,keepaspectratio]{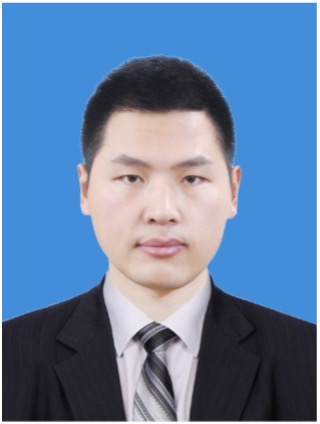}}]{Kejun Wu} (Senior Member, IEEE) received the Ph.D. degree in Information and Communication Engineering, Huazhong University of Science and Technology, Wuhan, China, in 2022. He has worked as a Research Fellow in the School of Electrical and Electronic Engineering, Nanyang Technological University, Singapore from 2022 to 2024. He is currently a Lecturer in the School of Electronic Information and Communications, Huazhong University of Science and Technology. His research interests include generative artificial intelligence, vision large language models, image restoration, and video compression, etc. He has published over 40 peer-reviewed papers including IEEE TMM, TCSVT, TOMM, OE, NeurIPS, etc. He has served as Session Chair in international conferences ICASSP 2024, IEEE ISCAS 2024, IEEE MMSP 2023, and Lead Guest Editor in JVCI.   
\end{IEEEbiography}

\begin{IEEEbiography}[{\includegraphics[width=1in,height=1.25in,clip,keepaspectratio]{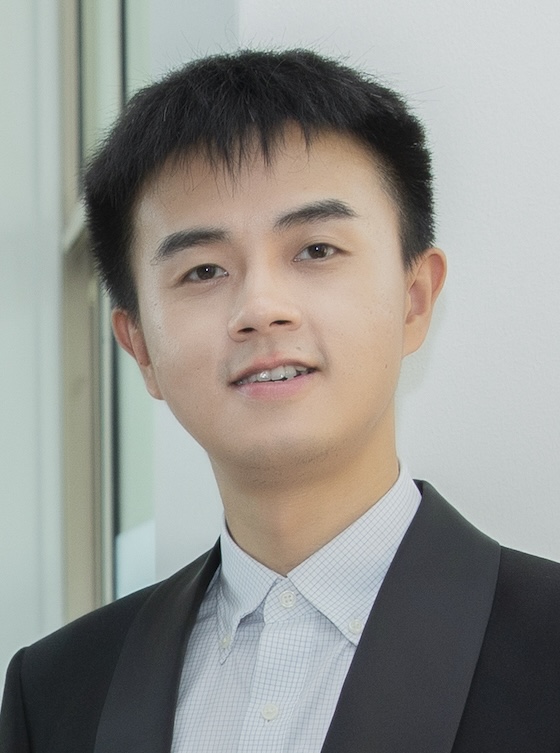}}]{Yi Wang}
(Member, IEEE) received BEng degree in electronic information engineering and MEng degree in information and signal processing from the School of Electronics and Information, Northwestern Polytechnical University, Xi’an, China, in 2013 and 2016, respectively. He earned PhD in the School of Electrical and Electronic Engineering from Nanyang Technological University, Singapore, in 2021. He is currently a Research Assistant Professor at the Department of Electrical and Electronic Engineering, The Hong Kong Polytechnic University, Hong Kong. His research interests include Image/Video Processing, Computer Vision, Intelligent Transport Systems, and Digital Forensics.
\end{IEEEbiography}

\begin{IEEEbiography}[{\includegraphics[width=1in,height=1.5in,clip,keepaspectratio]{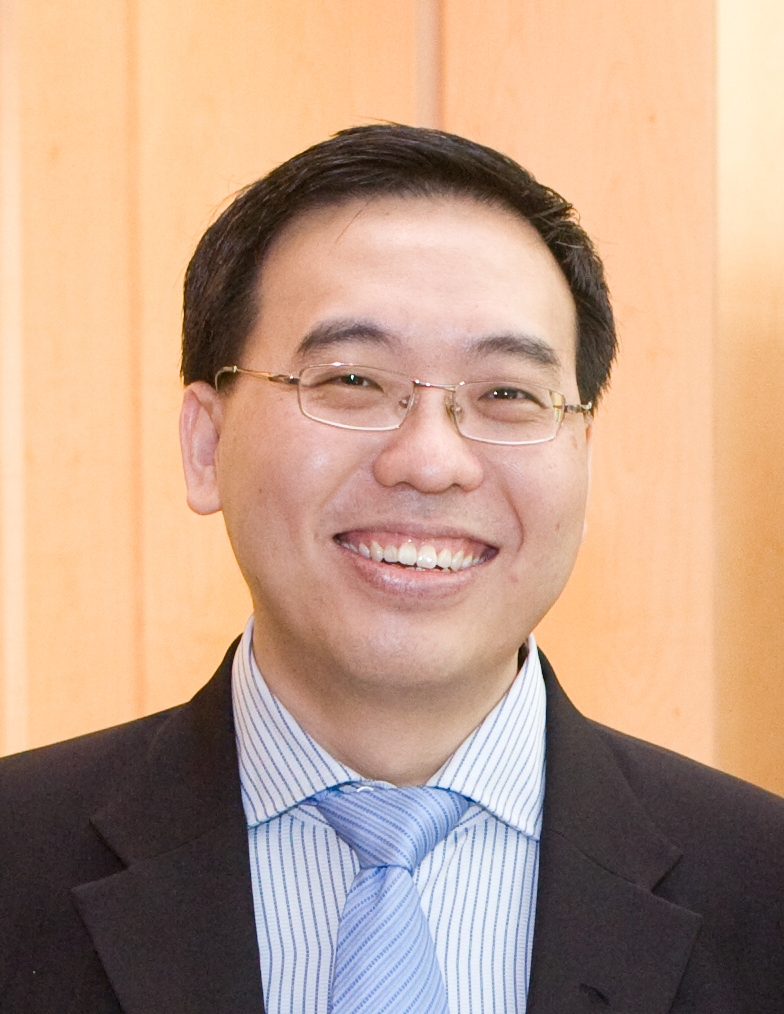}}]{Kim-Hui Yap} (Senior Member, IEEE) received the BEng and Ph.D. degrees in electrical engineering from the University of Sydney, Australia. He is currently an Associate Professor with the School of Electrical and Electronic Engineering, Nanyang Technological University, Singapore. He has authored more than 100 technical publications in various international peer-reviewed journals, conference proceedings, and book chapters. He has also authored a book titled \textit{Adaptive Image Processing: A Computational Intelligence Perspective} (Second Edition, CRCPress). His current research interests include artificial intelligence, data analytics, image/video processing, and computer vision. He has participated in the organization of various international conferences, serving in different capacities, including the technical program co-chair, the finance chair, and the publication chair in these conferences. He was an associate editor and an editorial board member of several international journals.
\end{IEEEbiography}

\begin{IEEEbiography}[{\includegraphics[width=1in,height=1.5in,clip,keepaspectratio]{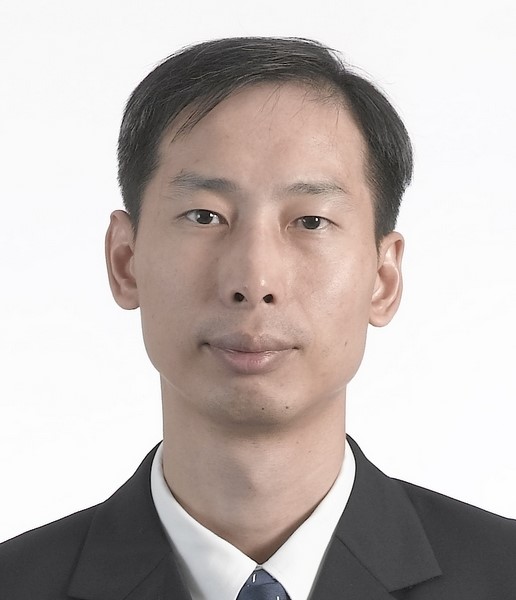}}]{Lap-Pui Chau}
(Fellow, IEEE) received a Ph.D. degree from The Hong Kong Polytechnic University in 1997. He was with the School of Electrical and Electronic Engineering, Nanyang Technological University from 1997 to 2022. He is currently a Professor in the Department of Electrical and Electronic Engineering, The Hong Kong Polytechnic University. His current research interests include image and video analytics, and autonomous driving. He is an IEEE Fellow. He was the chair of Technical Committee on Circuits Systems for Communications of IEEE Circuits and Systems Society from 2010 to 2012. He was general chairs and program chairs for some international conferences.
\end{IEEEbiography}

\end{document}